\documentclass[11pt]{article}

\usepackage[final]{acl}

\usepackage{times}
\usepackage{latexsym}

\usepackage[T1]{fontenc}
\usepackage[utf8]{inputenc}

\usepackage{microtype}

\usepackage{inconsolata}

\usepackage{graphicx}
\usepackage{algorithm}
\usepackage{algorithmic}
\usepackage{amssymb}
\usepackage{amsmath}
\usepackage{booktabs}
\usepackage{multirow}
\usepackage{adjustbox}
\usepackage{subcaption}
\usepackage{dsfont}
\usepackage{epsfig}
\usepackage{epstopdf}
\usepackage[table]{xcolor}
\usepackage{newfloat}
\usepackage{listings}
\usepackage{enumitem}
\DeclareCaptionStyle{ruled}{labelfont=normalfont,labelsep=colon,strut=off} % DO NOT CHANGE THIS
\floatstyle{ruled}
\newfloat{listing}{tb}{lst}{}

\floatname{listing}{Listing}
\usepackage[most,skins,theorems]{tcolorbox}
\tcbset{
  takeawaysbox/.style={
    title=Takeaways,
    colback=default!80,
    colframe=black,
    fonttitle=\bfseries\small,
    coltitle=white,
    colbacktitle=black,
    enhanced,
    attach boxed title to top left={xshift=2.5mm,yshift=-2.5mm},
    boxed title style={rounded corners, size=small, colframe=black, colback=black},
    width=\linewidth,
    arc=3.5mm
  }
}
\definecolor{lightblue}{RGB}{220,235,250}  % 浅蓝
\definecolor{lightgreen}{RGB}{230,242,222} % 淡绿
\definecolor{lightyellow}{RGB}{255,255,210} % 淡黄
\definecolor{lightorange}{RGB}{255,235,210} % 淡橙
\definecolor{default}{RGB}{255,255,255}
\definecolor{mgreen}{RGB}{6,128,67}
\definecolor{mgray}{RGB}{128,128,128}
\definecolor{mygreen}{RGB}{233,247,234}
\definecolor{mygray}{gray}{0.97}

\usepackage{amsmath,amssymb,amsthm}
\usepackage{mathtools}
\usepackage{natbib}
\usepackage{hyperref}

\theoremstyle{plain}
\newtheorem{theorem}{Theorem}[section]
\newtheorem{proposition}{Proposition}[section]
\newtheorem{lemma}[theorem]{Lemma}

\theoremstyle{remark}
\newtheorem*{remark}{Remark}

\NewDocumentCommand{\circnumpink}{O{pink!30!white} m}{%
  \tikz[baseline=(char.base)]{
    \node[shape=circle, fill=#1, inner sep=1pt, font=\small\bfseries] (char)
      {\textcolor{pink!70!red}{#2}};
  }%
}

\title{TemplateRL: Structured Template-Guided Reinforcement Learning for LLM Reasoning}

\author{
 \textbf{Jinyang Wu\textsuperscript{1}\thanks{\quad Equal Contribution}},
 \textbf{Chonghua Liao\textsuperscript{1}\footnotemark[1]},
 \textbf{Mingkuan Feng\textsuperscript{1}\footnotemark[1]},
 \textbf{Shuai Zhang\textsuperscript{1}},\\
 \textbf{Zhengqi Wen\textsuperscript{5}},
 \textbf{Haoran Luo\textsuperscript{2}},
 \textbf{Ling Yang\textsuperscript{3}},
 \textbf{Huazhe Xu\textsuperscript{1,4}},
 \textbf{Jianhua Tao\textsuperscript{1,5}}
\\
 \textsuperscript{1}Tsinghua University,
 \textsuperscript{2}Nanyang Technological University,\\
 \textsuperscript{3}Princeton University,
 \textsuperscript{4}Shanghai AI Lab,\\
 \textsuperscript{5}Beijing National Research Center for Information Science and Technology
\\
\texttt{\{wu-jy23,lch22\}@mails.tsinghua.edu.cn}
}

\begin{document}
\maketitle
\begin{abstract}
Reinforcement learning (RL) has emerged as an effective paradigm for enhancing model reasoning. However, existing RL methods like GRPO typically rely on unstructured self-sampling to fit scalar rewards, often producing inefficient rollouts that fail to capture transferable problem-solving strategies. To address this limitation, we propose \textbf{TemplateRL}, a structured template-guided RL framework that augments policy optimization with explicit template guidance. Our approach first constructs a problem-solving template library via MCTS on a small seed set, then seamlessly integrates this high-level structured guidance into RL training. By guiding rollout generation to align with proven template structures, TemplateRL significantly improves high-quality trajectory hit rates while reducing ineffective exploration. This structure-guided design steers the policy toward validated strategic patterns, stabilizing training dynamics, and enhancing RL sampling efficiency. Notably, the explicit template library is interpretable, editable, and supports online updates-enabling continuous updates during both training and inference. Extensive experiments demonstrate that TemplateRL outperforms GRPO by 99\% on AIME and 41\% on AMC, with superior stability on weak models and remarkable cross-domain generalization, highlighting its potential for broader tasks.
\end{abstract}

% -----------------------
% method part
% -----------------------
\section{Introduction}\label{section1}
Reinforcement learning (RL) has demonstrated remarkable success in enhancing the reasoning capabilities of large language models (LLMs), as exemplified by OpenAI-o1~\citep{jaech2024openai}, DeepSeek-R1~\citep{guo2025deepseek}, and Kimi-1.5~\citep{team2025kimi}. In contrast to traditional approaches that rely on human-curated annotations~\citep{achiam2023gpt,grattafiori2024llama}, contemporary RL training paradigms~\citep{shao2024deepseekmath} directly optimize base models using automatically computable reward signals. This enables models to develop sophisticated capabilities, including problem decomposition and self-reflection~\citep{gandhi2025cognitive,yue2025does}.

\begin{figure}[t]
  \centering
  \includegraphics[width=0.99\linewidth]{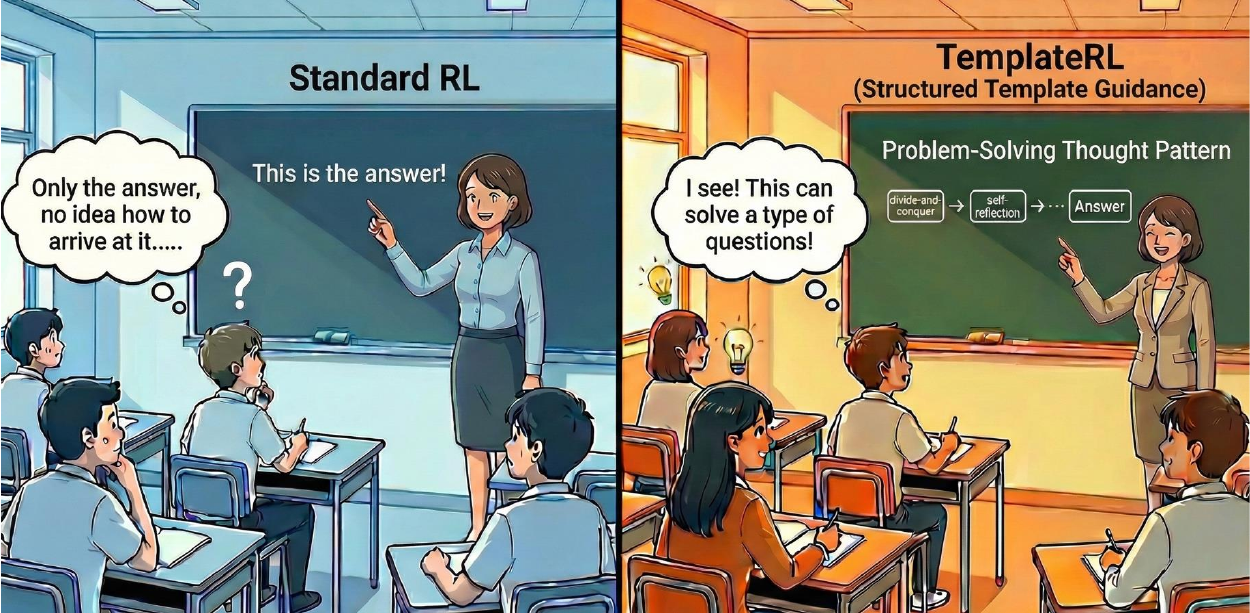}
    \caption{\textbf{Paradigm comparison: Teacher-Student Analogy for RL.} Standard RL like GRPO (\textbf{Left}) provides only sparse answer rewards, while TemplateRL (\textbf{Right}) offers structured templates encoding problem-solving thought patterns, enabling effective learning of both concrete steps and underlying strategic logic.}
  \label{story}
  \vspace{-0.2in}
\end{figure}

Recent RL research primarily focuses on three directions: (1) Algorithmic Refinements that address inherent limitations such as length bias, KL divergence constraints, and advantage estimation~\citep{yu2025dapo,liu2025understanding}; (2) Data-Level Optimizations that reduce annotation dependencies and improve data organization to enable minimally supervised training~\citep{zuo2025ttrl,wang2025reinforcement}; (3) Exploration Enhancement through entropy regularization or token-level learning to encourage distributional diversity~\citep{wang20258020rulehighentropyminority,cui2025entropy}.

While promising, they remain fundamentally constrained by unstructured self-sampling within the model's internal distribution, lacking mechanisms to leverage structured guidance for more effective policy learning. This leads to three critical limitations: \textbf{(a) Inefficient trajectory sampling:} without structured guidance, models struggle to consistently generate high-quality trajectories, resulting in low hit rates and training instability on weak models; \textbf{(b) Difficulty in learning transferable strategies:} existing RL methods tends to implicitely learn from surface-level steps rather than extract generalizable high-level thought patterns (e.g., divide-and-conquer), hindering knowledge transfer and accumulation across domains; \textbf{(c) Lack of interpretability:} existing RL optimization often produce reasoning trajectories without explicit strategic structure, hindering error diagnosis and expert intervention in critical applications.

To this end, we propose \textbf{TemplateRL}, a novel template-guided reinforcement learning framework that elegantly integrates structured thought templates into policy optimization (Figure~\ref{story}). Our approach consists of three parts: \textbf{(1) template construction:} We employ Monte Carlo Tree Search on a small seed set to explore diverse solutions, then abstract successful paths into structured templates encoding high-level problem-solving patterns; \textbf{(2) template-guided training:} During RL training, we adaptively retrieve relevant templates as structured guidance for rollout generation, steering the policy toward proven strategic patterns; \textbf{(3) optional template updates:} The template library supports continuous updates by automatically abstracting new patterns from successful RL rollouts. This design delivers three key advantages: For \textbf{(a)}, structured guidance improves solution hit rates and stabilizes training on weaker models; For \textbf{(b)}, explicit templates enable learning both concrete steps and underlying strategic logic, facilitating cross-domain generalization and flexible knowledge updates; For \textbf{(c)}, template-guided trajectories form interpretable decision chains that support error diagnosis and expert refinement. 

Extensive experiments validate TemplateRL's effectiveness across four dimensions: \textbf{(1) Improved Performance:} TemplateRL significantly outperforms GRPO by 99\% on AIME, and 41\% on AMC on a 7B backbone, surpassing representative baselines across benchmarks, model scales and architectures; \textbf{(2) Enhanced Training Stability:} Unlike GRPO which easily collapses on weaker models (e.g., Llama-3.2-3B), TemplateRL achieves stable training dynamics; \textbf{(3) Broad Generalization:} Our method demonstrates strong cross-domain performance on out-of-distribution benchmarks, including BALROG (agent), GPQA-D (science), and MMLU-Pro (knowledge). It consistently improves across different modalities (text, vision); \textbf{(4) Dynamic Extensibility:} Template library supports continuous updates during both training and inference, enabling incremental strategy incorporation. Further analysis reveal that TemplateRL produces clear and interpretable reasoning trajectories.

\begin{figure*}[htbp!]
\begin{center}
\centerline{\includegraphics[width=0.99\textwidth]{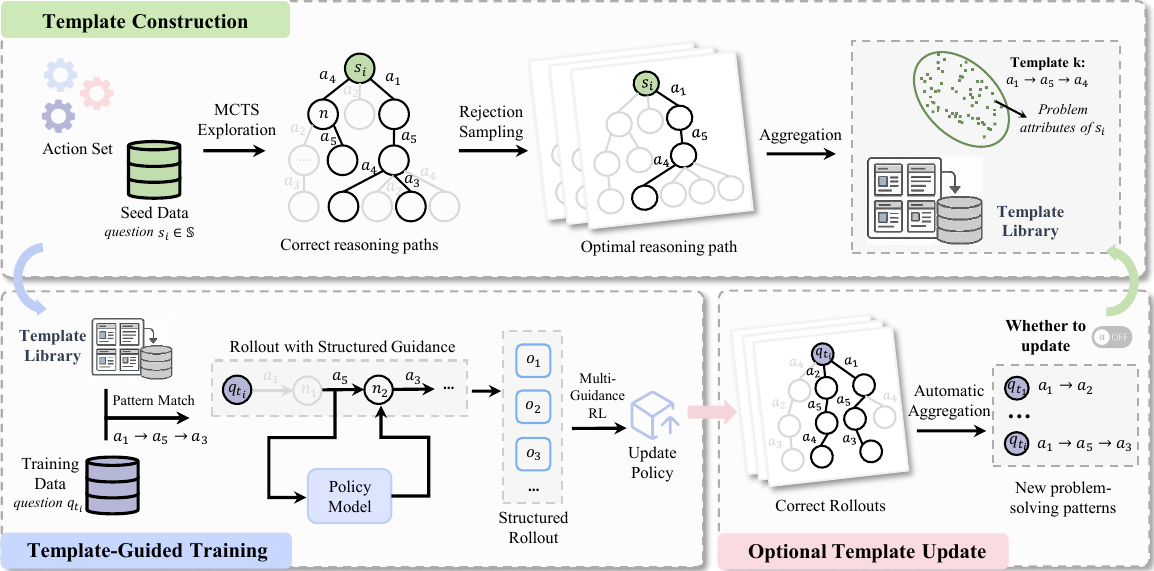}}
\caption{\textbf{Flowchart of TemplateRL.} This framework consists of three components: (1) template construction (Section~\ref{section2.3}); (2) template-guided training (Section~\ref{section2.2}); and (3) optional template updates (Section~\ref{updates}).}
\label{Figure2}
\end{center}
\vskip -0.25in
\end{figure*}

\section{Preliminaries}
\paragraph{LLMs as Markov Decision Processes.}
LLM generation can be formulated as a token-level Markov Decision Process $\mathcal{M}(\mathcal{S}, \mathcal{A}, r, p_{\mathcal{Q}})$~\citep{sutton1998reinforcement,liu2025understanding}, where $\mathcal{S}$ represents states (observation sequences) and $\mathcal{A}$ denotes the action space (vocabulary). At each step $t$, state $s_t \in \mathcal{S}$ consists of the input question $\mathbf{q}$ concatenated with all tokens generated so far $\mathbf{o}_{<t}$. This state serves as input to policy model $\pi_\theta(\cdot | s_t)$. Specifically, the policy processes $s_t=(\mathbf{q},\mathbf{o}_{<t})=(q_1,q_2,\dots,q_l,o_1,o_2,\dots,o_{t-1})$, where $q_i$ denotes the $i$-th token of question $\mathbf{q}$ and $o_{j,<t}$ represents the token generated by $\pi_\theta$ at step $j$. In the RL framework, the policy then samples the next token from $\mathcal{A}$. The entropy-regularized optimization objective~\citep{schulman2017equivalence} is:
\begin{equation}
\begin{aligned}
\label{eq:rl_objective}
    \mathcal{J}(\pi_\theta) = \underset{{\mathbf{q} \sim p_{\mathcal{Q}}}}{\mathbb{E}}[ \underset{\mathbf{o} \sim \pi_\theta(\cdot|\mathbf{q})}{\mathbb{E}}[R(\mathbf{q}, \mathbf{o})] - \\
    \beta \cdot \mathbb{D}_{KL}[\pi_\theta(\cdot|\mathbf{q})) || \pi_{\text{ref}}(\cdot|\mathbf{q})]]\,,
\end{aligned}
\end{equation}
where $R(\mathbf{q}, \mathbf{o})=\sum_{t=1}^{|\mathbf{o}|}r(s_t, o_t)$ is the return of trajectory $(\mathbf{q};\mathbf{o})$, $r(\cdot)$ is the reward model, and $\pi_{\text{ref}}$ is a reference policy. The KL regularization term prevents $\pi_\theta$ from deviating excessively from $\pi_{\text{ref}}$.

\paragraph{Template Action.}
A template action is a prompt \( a \) designed to elicit a specific type of reasoning from a model, given a partial solution as input. It encapsulates an abstract human-reasoning procedure, such as proposing the next sub-question or generating the subsequent inference step~\citep{guan2025rstar}. More precisely, given a partial solution $\mathbf{p}$, \( a \) induces the policy model $\pi_\theta$ to generate the next partial solution $\mathbf{p}'$:
\begin{equation}
\begin{aligned}
\mathbf{p}'=\pi_\theta(a(\mathbf{p}))\,.
\end{aligned}
\end{equation}

\section{Template-Guided RL}\label{section2}
As shown in Figure~\ref{Figure2}, TemplateRL consists of three parts: (1) template construction (Section~\ref{section2.3}); (2) template-guided training (Section~\ref{section2.2}); and (3) optional template updates (Section~\ref{updates}).

% \subsection{Template Library and Augmented Reasoning}\label{section2.3}
\subsection{Template Construction}\label{section2.3}
We first describe how to construct a template library, which guides RL training in Section~\ref{section2.2}. Previous work~\citep{Kahneman2011} reveals that humans solve complex reasoning tasks by applying universal guidelines (``templates'') induced from similar problems rather than starting from scratch. These high-level templates help address unfamiliar tasks by leveraging previously successful strategies. Inspired by this, we introduce ``template library'', a lightweight hub of structured templates that provide high-level strategic guidance for RL.

\begin{figure}[t!]
\begin{center}
    \includegraphics[width=0.7\columnwidth]{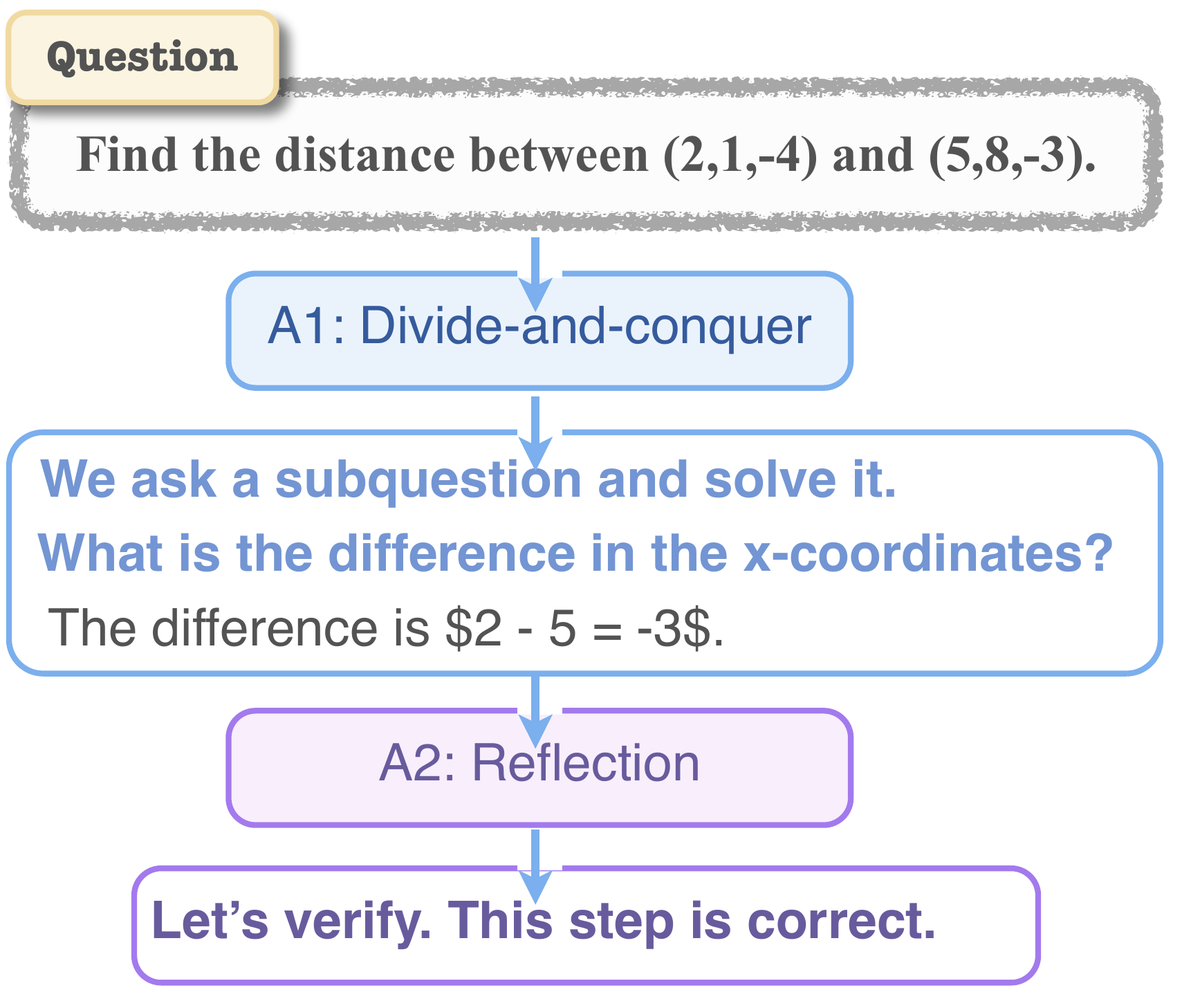}
    \caption{Structure of action-chain solution trajectories.}
    \label{fig:method}
\end{center}
\vskip -0.2in
\end{figure}

% \paragraph{Template Library.}
\paragraph{\circnumpink{1} MCTS Exploration.}
Starting with a small seed set $\mathcal{S}=\left\{\mathbf{s}_1, \dots,\mathbf{s}_{s}\right\}$, we employ Monte Carlo Tree Search (MCTS)~\citep{kocsis2006bandit} to generate solution trees. For each question $\mathbf{s}_i \in \mathcal{S}$, given a predefined action set $\mathcal{A}=\{a_1,\dots,a_{|\mathcal{A}|}\}$ and model $\pi_\theta$, MCTS build a search tree $\mathcal{T}_i$ where: the root node represents question $\mathbf{s}_i$, each edge denotes an action $a\in \mathcal{A}$, and each child node $n$ contains partial solutions generated by $\pi_\theta$ under the corresponding action. A path from root $\mathbf{s}_i$ to leaf node $\mathbf{n}_{i,j,d_{i,j}}$ forms a solution trajectory:
\begin{equation}
    \mathbf{t}_{i,j}=\big(\mathbf{s}_i, a_{i,j,1}, \mathbf{n}_{i,j,1}, \dots, a_{i,j,d_{i,j}}, \mathbf{n}_{i,j,d_{i,j}}\big),
\end{equation}
Each intermediate node $\mathbf{n}_{i,j,\ell}$ is generated based on the cumulative context of its parent nodes and current action: 
\begin{equation}
\mathbf{n}_{i,j,\ell}=\pi_\theta([\mathbf{s}_i, a_{i,j,1}, \mathbf{n}_{i,j,1}, \dots,a_{i,j,\ell}]).
\end{equation}
Through this process, for each seed question $\mathbf{s}_i$, we obtain multiple solution traces $\mathbb{T}_i=\{\mathbf{t}_{i,1},\dots,\mathbf{t}_{i,|\mathbb{T}_i|}\}$ (e.g., Figure~\ref{fig:method}). The MCTS algorithm assigns a final reward $R(\mathbf{t}_{i,j}\mid \mathbf{s}_i)$ to each trace $\mathbf{t}_{i,j}\in\mathbb{T}_i$. More details are in Appendix~\ref{B}.
% \vskip-0.1in

\vskip-0.3in
\paragraph{\circnumpink{2} Rejection Sampling.}
After MCTS exploration, we obtain multiple solution traces for each seed question $\mathbf{s}_i$. To identify the optimal path, we use a simple path selection metric from~\citet{wu2024beyond}:
\begin{equation}
\text{Score}(\mathbf{s}_i,\mathbf{t}_{i,j}) = b \cdot R(\mathbf{t}_{i,j}\mid \mathbf{s}_i) - (1-b)\cdot C(\mathbf{t}_{i,j}),
\label{eq:voc_score}
\end{equation}
where $C(\mathbf{t}_{i,j})$ denotes path complexity (e.g, action count, $C(\mathbf{t}_{i,j})=d_{i,j}$), and $b\in[0,1]$ balances solution quality and complexity. This scoring function selects paths that maximize accuracy while maintaining procedural conciseness. For each question $\mathbf{s}_i\in\mathcal{S}$, we select the optimal trace via:
\begin{equation}
\mathbf{t}_{i,\text{best}}=\arg\max_{\mathbf{t}_{i,j}\in\mathbb{T}_i}\ \text{Score}(\mathbf{s}_i,\mathbf{t}_{i,j}).
\end{equation}
\vskip -1.5in
\paragraph{\circnumpink{3} Aggregation.}
Since each node in $\mathbf{t}_{i,\text{best}}$ corresponds to an instantiated action $a_{i,\ell}\in\mathcal{A}$, we retain the action sequence as a high-level thought template $T_i=(a_{i,1},\dots,a_{i,d_{i,\text{best}}})$. We then aggregate identical patterns to construct a template library $\mathcal{L}=\{\hat T_1,\dots,\hat T_{|\mathcal{L}|}\}$. Aggregation is guided by Problem Condition Complexity (PCC)~\citep{lee2000problem}, defined as the number of prior conditions in $\mathbf{s}_i$; we obtain $\mathrm{PCC}(\mathbf{s}_i)\in\mathbb{R}_{\ge 0}$ by prompting $\pi_\theta$ with $\mathbf{s}_i$. Each template stores both a structured thought pattern (e.g., $a_1\rightarrow a_2$) and the average PCC of questions sharing this pattern:
\begin{equation}
    \hat T_j=\big(\mathrm{PCC}_{T_j},\, T_j\big), \mathrm{PCC}_{T_j}=\frac{1}{|\mathcal{I}_j|}\sum_{i\in\mathcal{I}_j}\mathrm{PCC}(\mathbf{s}_i).
\end{equation}
where $\mathcal{I}_j=\{i:\,T_i=T_j\}$. These templates represent generalized problem-solving strategies and serve as structured guidance in Section~\ref{section2.2}. More details are in Appendix~\ref{B}.

Notably, while we use MCTS to build reasoning trees and abstract them into structured templates, such templates can also be obtained through other ways, such as expert-crafted general solutions for problem categories. We leave this for future work.

\subsection{Template-Guided Training}\label{section2.2}
Taking GRPO as an example, we now integrate high-level structured template guidance into RL to enhance policy learning. We exploit the observation that for multi-step reasoning, models are more likely to generate one correct step than to complete the entire reasoning steps in a single inference.

\vskip -0.1in
\paragraph{Rollout with Structured Guidance.}
Following meta-reasoning principles~\citep{russell1991principles}, we adaptively retrieve relevant templates for each training question $\mathbf{q}_{}$. We compute its $\mathrm{PCC}(\mathbf{q}_{})$, define the distance $d(\mathbf{q},\hat T_j)=\lvert \mathrm{PCC}(\mathbf{q}_{})-\mathrm{PCC}_{T_j}\rvert$ to each template $\hat{T}_j\in\mathcal{L}$, and select the top-$k$ most similar $\{\hat T_{i_1},\dots,\hat T_{i_k}\}$.

Given policy $\pi_\theta$, query $\mathbf{q}$, and a retrieved template $T_{i}=(a_1,\dots,a_{d_i})$, we roll out iteratively. Let $\oplus$ denote sequence concatenation. Define the history
$\mathbf{p}_0=\mathbf{q}$ and for $\ell=1,\dots,d_i$,
\begin{equation}
\mathbf{s}_{\ell}\sim \pi_\theta\big(\cdot \,\big|\, \mathbf{p}_{\ell-1}, a_\ell\big),
\mathbf{p}_{\ell}=\mathbf{p}_{\ell-1}\oplus a_\ell\oplus \mathbf{s}_{\ell}.
\end{equation}
The final template-guided rollout is $\mathbf{o}=\mathbf{p}_{d_i}$.

\paragraph{Multi-Guidance RL.}
Taking GRPO as an example, we now investigate how multiple template guidance decomposes RL's learning objective into templated sub-objectives, and elucidate how templates enhance sampling quality and RL training.

As illustrated above, given diverse templates $\{T_1,\dots,T_{|g|}\}$, we independently sample $G_i$ rollouts $\{\mathbf{o}_{i,1},\dots,\mathbf{o}_{i,G_i}\}$ for each template $T_i$ and use all rollouts to update the policy. Omitting KL term for simplicity, the optimization objective becomes
\begin{equation}
\begin{gathered}
\tilde{\mathcal{J}}_{\text{GRPO}}(\pi_{\theta}) = \frac{1}{\sum_{i=1}^{|g|} G_i} \sum_{i=1}^{|g|}  \sum_{j=1}^{G_i}\frac{1}{|\mathbf{o}_{i,j}|}\sum_{t=1}^{|\mathbf{o}_{i,j}|}\\ \min\left[\rho_{i,j,t}{A}_{i,t}, \hat{\rho}_{i,j,t}{A}_{i,t}\right] \,, \\%+ 
\end{gathered}
\label{eq:extendgrpo}
\end{equation}
where the probability ratio is
\(\rho_{i,j,t} = \frac{\pi_\theta(o_{i,j,t} | {\mathbf{q}},\mathbf{o}_{i,j,<t})}{\pi_{\theta_{\text{old}}}(o_{i,j,t} | {\mathbf{q}},\mathbf{o}_{i,j,<t})}\),
and the clipped ratio is
\(\hat{\rho}_{i,j,t}=\text{clip}(\rho_{i,j,t}; 1-\epsilon, 1+\epsilon)\).

For convenience, define the template-specific objective as:
\begin{equation}
\begin{aligned}
\mathcal{J}_{i}(\pi_{\theta}) = \frac{1}{G_i}  \sum_{j=1}^{G_i}\frac{1}{|\mathbf{o}_i|}&\sum_{t=1}^{|\mathbf{o}_i|} \min\left[\rho_{i,j,t}{A}_{i,t}, \hat{\rho}_{i,j,t}{A}_{i,t}\right] \,,
\end{aligned}
\end{equation}
Then, Equation \eqref{eq:extendgrpo} can be rewritten as
% \vskip-0.1in
\begin{equation}
\begin{gathered}
\tilde{\mathcal{J}}_{\text{GRPO}}(\pi_{\theta}) = \frac{1}{\sum_{i=1}^{|g|} G_i} \sum_{i=1}^{|g|}  G_i \mathcal{J}_{i}(\pi_{\theta})\,.
\end{gathered}
\label{eq:simple_extendgrpo}
\end{equation} 
which shows that template-guided training optimizes the model across multiple structured strategic patterns, with $G_i$ weighting each template's contribution. We provide theoretical analysis below, and detailed proofs in Appendix~\ref{app:theory_grpo_seed_transfer}.

\begin{table*}[t]
\centering
% 移除可能导致溢出的缩放，改用固定列宽（更稳定）
\resizebox{1.0\textwidth}{!}{
\begin{tabular}{lccccccc}
\toprule
\textbf{Model}  & \textbf{MATH500 $\uparrow$} & \textbf{AIME24 $\uparrow$} & \textbf{AMC $\uparrow$} & \textbf{Minerva $\uparrow$} & \textbf{Olympiad $\uparrow$} &  \textbf{Avg. $\uparrow$} \\
\midrule
Qwen2.5-Math-7B-Base~\citep{qwen2.5_math}    & 50.8 & 13.3 & 42.5 & 12.1 & 17.2 & 27.2 \\
Qwen2.5-Math-7B-Ins.~\citep{qwen2.5_math}  & 81.0 & 13.3 & 55.0 & 32.7 & 38.8 & 44.1 \\
\midrule
SimpleRL-Zero~\citep{zeng2025simplerl}   & 74.6 & 26.7 & 60.0 & 27.6 & 35.8 & 44.9 \\
OpenReasoner-Zero~\citep{orz}  & 81.0 & 16.7 & 57.5 & 32.7 & 43.2 & 46.2 \\
PRIME-Zero~\citep{prime}   & 79.0 & 20.0 & 60.0 & 36.4 & 40.6 & 47.2 \\
Oat-Zero~\citep{liu2025understanding}    & 79.6 & 30.0 & 60.0 & 34.2 & 39.9 & 48.7 \\
\midrule
RLOO~\citep{ahmadian2024rloo} & 73.8 & 30.0 & 50.0 & 35.5 & 36.0 & 45.1 \\
\rowcolor[RGB]{236,244,252}
TemplateRL (Ours) & \textbf{78.8} & \textbf{33.3} & \textbf{67.5} & \textbf{36.7} & \textbf{41.2} & \textbf{51.5}$\ast$ \\
\rowcolor{pink!20}
\textbf{$\bigtriangleup$ $(\uparrow)$} & +6.8$\%$ & +11.0$\%$ & +35.0$\%$ & +3.4$\%$ & +14.4$\%$ & +14.2$\%$\\
\midrule
GRPO~\citep{shao2024deepseekmath}  & 76.2 & 16.7 & 55.0 & 32.7 & 38.1 & 43.8 \\
\rowcolor[RGB]{236,244,252}
TemplateRL (Ours)  & \textbf{83.4} & \textbf{33.3} & \textbf{77.5} & \textbf{38.2} & \textbf{46.2} & \textbf{55.8}$\ast$ \\
\rowcolor{pink!15}
\textbf{$\bigtriangleup$ ($\uparrow$)} & +9.4$\%$ & +99.4$\%$ & +40.9$\%$ & +16.8$\%$ & +21.2$\%$ & +27.4$\%$ \\
\bottomrule
\end{tabular}}
\caption{\textbf{Performance comparison on Qwen2.5-Math-7B-Base}. The best results on each benchmark are highlighted in \textbf{bold}. $\ast$ represents significantly better than baselines (p<0.05).}
\label{tab:main_results}
\end{table*}

\begin{proposition}[Stability and positive-sample guarantee. Proof in Appendix~\ref{thm:grouped_sampling}]
\label{thm:stability}
Consider guided GRPO with $|g|$ independent guidance groups, each containing $G_i = G/|g|$ trajectories. If the per-trajectory positive-advantage probability in template group $i$ is $p_i$, then the probability of at least one positive trajectory is
\begin{equation}
P_{\mathrm{pos}} = 1 - P_0 = 1 - \prod_{i=1}^{|g|} (1 - p_i)^{G_i}.
\end{equation}
Since $\log(1-p)$ is concave, we have $\prod_i (1-p_i)^{G_i} \le (1-\bar{p})^{G}$ for $\bar{p}=\frac{1}{|g|}\sum_i p_i$. Hence, grouping increases $P_{\mathrm{pos}}$ and reduces gradient variance, leading to improved training stability compared with the unguided setting.
\end{proposition}

\begin{proposition}[Template transfer improves success probability. Proof in Appendix~\ref{thm:grouped_template_transfer}]
\label{thm:transfer}
Let $T^\star(\mathbf{q}')$ denote a template obtained from MCTS for a seed question $\mathbf{q}'$ with positive advantage. 
If the advantage function $A(\tau;\mathbf{q})$ is $L$-Lipschitz continuous in $\mathbf{q}$
\begin{equation}
    A(T^\star(\mathbf{q}'); \mathbf{q})\ge A(T^\star(\mathbf{q}'); \mathbf{q}')-Ld(q, q')\,,
\end{equation}
then for any new query $\mathbf{q}$ satisfying
\begin{equation}
A(T^\star(\mathbf{q}'); \mathbf{q}') > L \, d(\mathbf{q}, \mathbf{q}')\,,
\end{equation}
we have $A(T^\star(\mathbf{q}'); \mathbf{q}) > 0$. 
Therefore, retrieving a high-quality template from similar questions yields a mini-group on $\mathbf{q}$, with positive advantage probability $r_h > p_h^{\mathrm{policy}}$ (random rollouts).
\end{proposition}

\subsection{Optional Template Update}\label{updates}
During RL training, we optionally leverage correct rollouts to continually enrich the template library. Specifically, given a correct rollout $\mathbf{o}$, we apply keyword-based or lightweight model-based pattern extraction to automatically identify the underlying action sequence (template) $T' = (a'_1,\dots,a'_d)$. Such templates are then incorporated into the library for continual expansion. We empirically validate the effectiveness of dynamic template updates during both training and inference in Section~\ref{Extensibility}.

\begin{figure*}[ht!]  
  \centering
  \vskip -0.1in
  \includegraphics[width=1.0\linewidth]{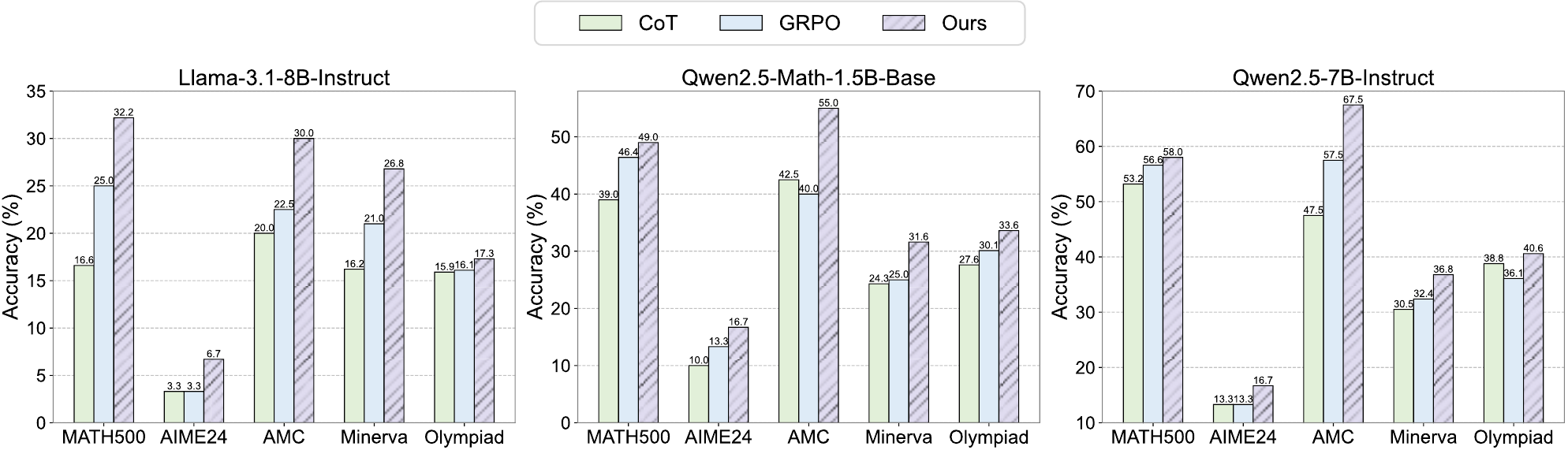}
    \caption{\textbf{Performance comparison across different model scales and architectures.} For better visualization, MATH500 results are adjusted by subtracting 20 points for all models.}
  \label{model_scales}
  % \vskip -0.15in
  \vskip -0.1in
\end{figure*}

\section{Experiments}\label{exps}
This section presents experimental settings, results, and analysis. We analyze TemplateRL from five aspects: (1) improved performance, (2) enhanced training stability, (3) broad generalization, (4) dynamic extensibility, and (5) ablation study.

\subsection{Experimental Setup}\label{Experimental Setup}
\paragraph{Training Datasets.} We use training data exclusively from MATH~\citep{hendrycks2021measuring} level 3-5 problems, yielding 5.5K examples. We randomly sample 500 instances to construct the template library, with the remaining 5K for training.

\paragraph{Evaluation.}
We evaluate on widely used reasoning benchmarks, including MATH500~\citep{hendrycks2021measuring}, AIME 2024~\citep{li2024numinamath}, AMC~\citep{li2024numinamath}, Minerva~\citep{dataset_minerva}, OlympiadBench~\citep{dataset_olympiad}, GSM8K~\citep{cobbe2021training}, College Math~\citep{tang2024mathscale}, and Gaokao23~\citep{zhang2023evaluating}. Since our training data focuses on math, we further assess out-of-domain generalization on BALROG~\citep{paglieri2025balrog} for agent tasks, GPQA-D~\citep{gpqa} for graduate-level science, and MMLU-Pro~\citep{mmlu_pro} for general knowledge answering.

\paragraph{Baseline Methods.} We benchmark TemplateRL with representative baselines, such as \textit{Standard GRPO}~\citep{shao2024deepseekmath} and \textit{Oat-Zero}~\citep{liu2025understanding}. More details are provided in Appendix~\ref{C}.

\paragraph{Implementation Details.}
Following previous work~\cite{prime,liu2025understanding}, we use Qwen2.5-Math-7B-Base~\cite{qwen2.5_math} as the default model. For RL training, we set $\beta=0$ (no KL loss) and employ Dr.GRPO loss~\cite{liu2025understanding}. Our configuration includes batch size 128, 16 samples per question, and 2 guidance templates ($|g|=2$). The reward is binary accuracy verified by Math-Verify. We train for 500 steps on 8 A100 GPUs. More details are provided in Appendix~\ref{C}.
h
\subsection{Improved Performance}\label{performance enhancement}
\paragraph{Main Results.}
Table \ref{tab:main_results} presents performance comparisons across five competition-level reasoning benchmarks. The results reveal three findings:

\begin{figure}[ht!]
\centering
\begin{subfigure}{0.47\linewidth}
  \centering
  \includegraphics[width=\linewidth]{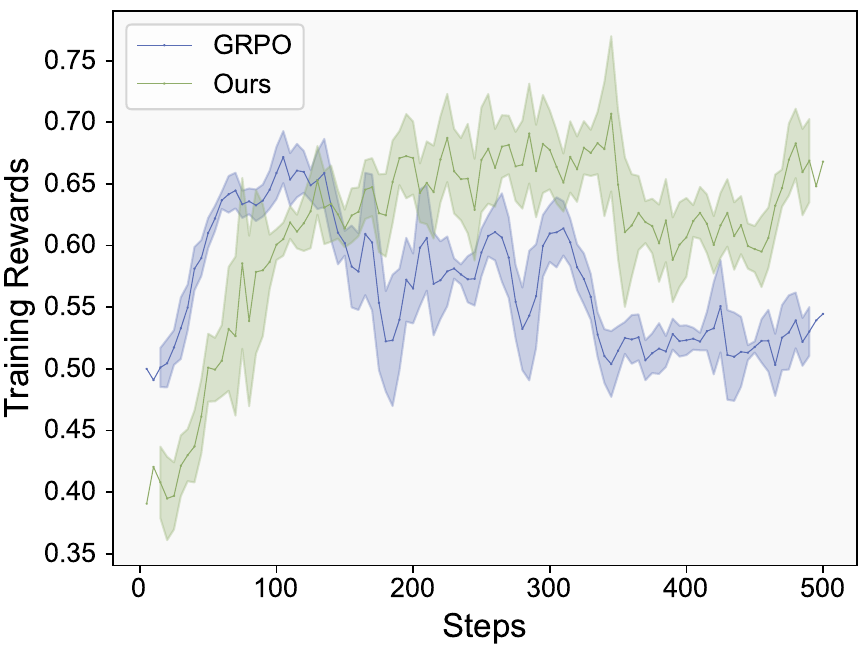}
  \caption{Qwen2.5-Math-7B-Base}
  \label{fig:7b}
\end{subfigure}
\hfill
\begin{subfigure}{0.47\linewidth}
  \centering
  \includegraphics[width=\linewidth]{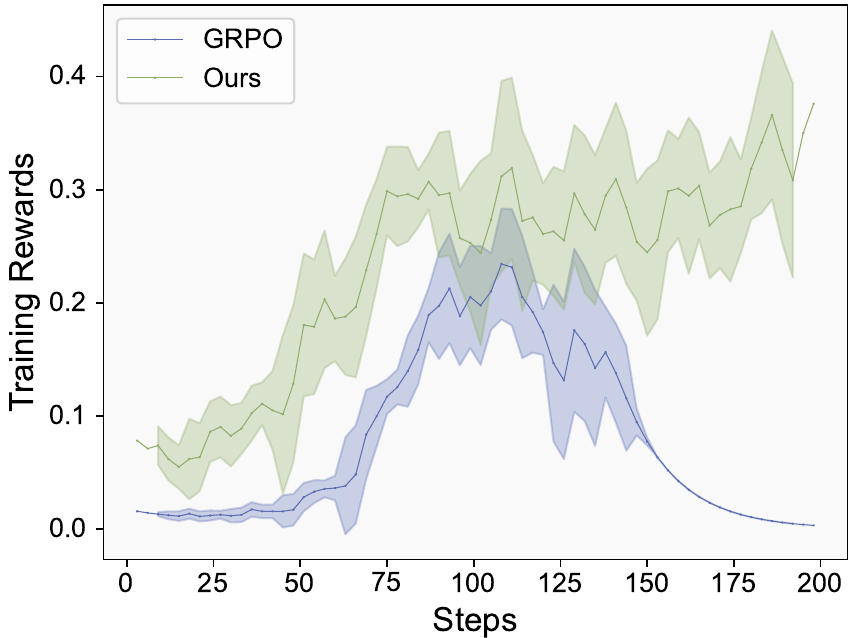}
  \caption{Llama-3.2-3B-Base}
  \label{fig:3b}
\end{subfigure}
\caption{\textbf{Training stability verification.} We evaluate on Qwen2.5-Math-7B-Base and Llama-3.2-3B-Base.}
\label{fig:reward_curve}
\vskip -0.18in
\end{figure}

$\diamondsuit$ \textit{TemplateRL Significantly Outperforms Baselines.} TemplateRL achieves an average score of 55.8, outperforming the best baseline (Oat-Zero) by 7.1 points and GRPO by 12.0 points. On AIME24, TemplateRL achieves 33.3\% versus GRPO's 16.7\% (+99.4\% relative gain), demonstrating the effectiveness of structure-guided policy optimization.

$\diamondsuit$ \textit{Greater Improvements on More Challenging Benchmarks.} Performance gains scale with task difficulty: MATH500 shows +9.4\% improvement, while more challenging AMC and AIME24 exhibit +40.9\% and +99.4\%, respectively. This suggests explicit template provide greater benefits for problems requiring complex strategic decomposition.

$\diamondsuit$ \textit{Structure-Guided Training Generalizes Across RL Algorithms.} When applied to other RL algorithms (e.g., RLOO), TemplateRL maintains substantial improvements. This consistency further confirms that structure-guided training is a general apporach applicable to diverse RL algorithms.

\vskip -0.2in

\paragraph{Scalability Across Different Models.}
As shown in Figure~\ref{model_scales}, TemplateRL maintains consistent improvements across: (1) different scales, from 1.5B to 8B parameters; (2) diverse architectures, both Qwen and Llama families show substantial improvements; and (3) various model types, both base models and instruction-tuned models benefit from structure-guided training (details in Table~\ref{table:moremodels_appendix}). We further validate template guidance on larger models (32B) in Appendix~\ref{capable_models}, demonstrating that structured guidance enhances rather than restricts exploration even for highly capable models.

\vskip -0.1in
\subsection{Enhanced Training Stability}\label{training dynamics}%
We analyze training stability by comparing reward curves on Qwen2.5-Math-7B-Base and Llama-3.2-3B-Base. As shown in Figure~\ref{fig:reward_curve}, TemplateRL achieves higher training rewards than GRPO on both models overall. Critically, GRPO exhibits training collapse after 100 steps on Llama-3.2-3B-Base, with rewards dropping from 0.3 to near 0.0. This aligns with prior findings~\citep{gandhi2025cognitive,liu2025understanding} that GRPO frequently fails on weaker models. In contrast, TemplateRL maintains stable dynamics, with rewards above 0.25.

TemplateRL's enhanced stability mainly stems from its structured template guidance during RL sampling. By providing explicit strategic patterns, templates equips weaker models with advanced decomposition capabilities typically exclusive to stronger models. This enables effective learning from challenging examples, which would otherwise yield sparse positive signal under standard GRPO training, as detailed in Section~\ref{section2}.

\begin{table*}[t!]
\centering
\vskip -0.1in
\resizebox{1.0\linewidth}{!}{
\begin{tabular}{lcccccc} 
    \toprule
    Method & MathVision~$\uparrow$ & MathVerse~$\uparrow$ & MathVista~$\uparrow$ & MMMU~$\uparrow$ & BLINK~$\uparrow$ & Avg.~$\uparrow$ \\
    \midrule
    CoT & 27.7 & 27.3 & 61.1 & 51.2 & 45.1 & 42.5 \\
    GRPO & 30.9 & 31.7 & 65.6 & 52.7 & 46.4 & 45.5 \\
    \rowcolor[RGB]{236,244,252}
    TemplateRL (Ours) & \textbf{34.6} & \textbf{36.4} & \textbf{70.2} & \textbf{56.8} & \textbf{48.3} & \textbf{49.3} \\
    \rowcolor{pink!15}
    \textbf{$\bigtriangleup$ $(\uparrow)$} & +12.0\% & +14.8\% & +7.0\% & +7.8\% & +4.1\% & +8.4\% \\
    \bottomrule
\end{tabular}
}
\caption{\textbf{Multimodal Extension.} We extend TemplateRL to multimodal domains with Qwen2.5-VL-3B-Instruct, and evaluate on multimodal reasoning benchmarks. TemplateRL consistently improves over GRPO across mathematical reasoning (MathVision, MathVerse, MathVista), general understanding (MMMU), and visual perception (BLINK).}\label{multimodal_results}
\vskip -0.08in
\end{table*}

\subsection{Broad Generalization}\label{broad generalization}
\begin{figure}[ht!]  
  \centering
  \includegraphics[width=1.0\linewidth]{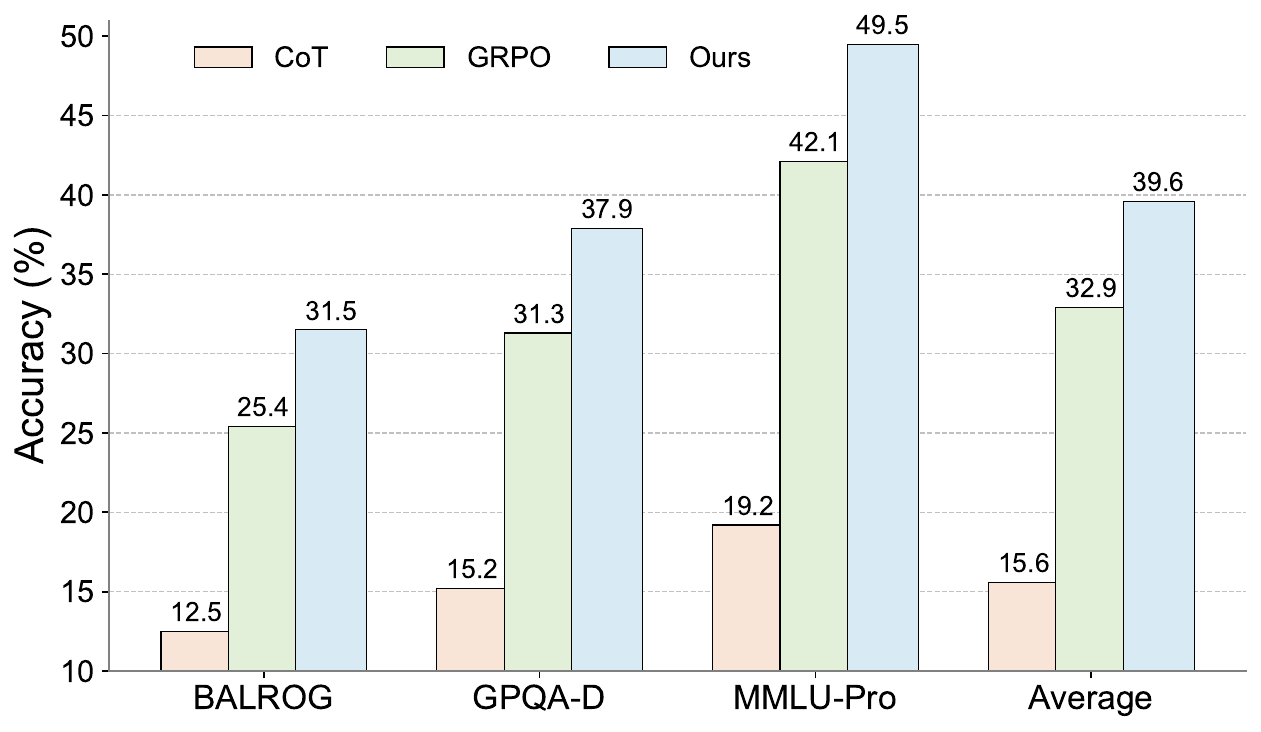}
  \caption{\textbf{Cross-domain generalization verification.} We provide OOD results on Qwen2.5-Math-7B-Base.}
  \label{ood_figure}
  \vskip -0.1in
\end{figure}

As shown in Table~\ref{multimodal_results} and Figure~\ref{ood_figure}, we examine TemplateRL's broad generalization through out-of-domain evaluation and multimodal extensions.

\paragraph{Cross-Domain Generalization.} 
To assess out-of-domain (OOD) performance, we evaluate TemplateRL on diverse reasoning tasks beyond mathematical domains: BALROG (agentic reasoning in games), GPQA-D (graduate-level science), and MMLU-Pro (general knowledge). As shown in Figure~\ref{ood_figure}, TemplateRL consistently outperforms GRPO across OOD tasks, with +6.1\% performance gains on complex agentic scenarios (BALROG). This reveals that high-level template guidance effectively enhances model generalization to practical applications. Detailed results are in Appendix~\ref{D.1}.

\paragraph{Multimodal Extension.}
Beyond textual reasoning, we extend TemplateRL to multimodal domains. We use the Geometry3K dataset~\citep{lu-etal-2021-inter} to construct MMTrain10K as training data and evaluate on diverse benchmarks, including mathematical reasoning (MathVision, MathVerse, MathVista), general understanding (MMMU), and visual perception (BLINK). Our settings are: batch size 128, 16 samples per question, and 2 template guidance. As shown in Table~\ref{multimodal_results}, TemplateRL consistently outperforms GRPO across all benchmarks, achieving +8.4\% average improvement. This indicates that structured template guidance effectively generalizes beyond text modality to multimodal scenarios. More details are provided in Appendix~\ref{D.3}.

\begin{figure*}[t!]
\vskip -0.1in
\begin{center}
\centerline{\includegraphics[width=0.96\textwidth]{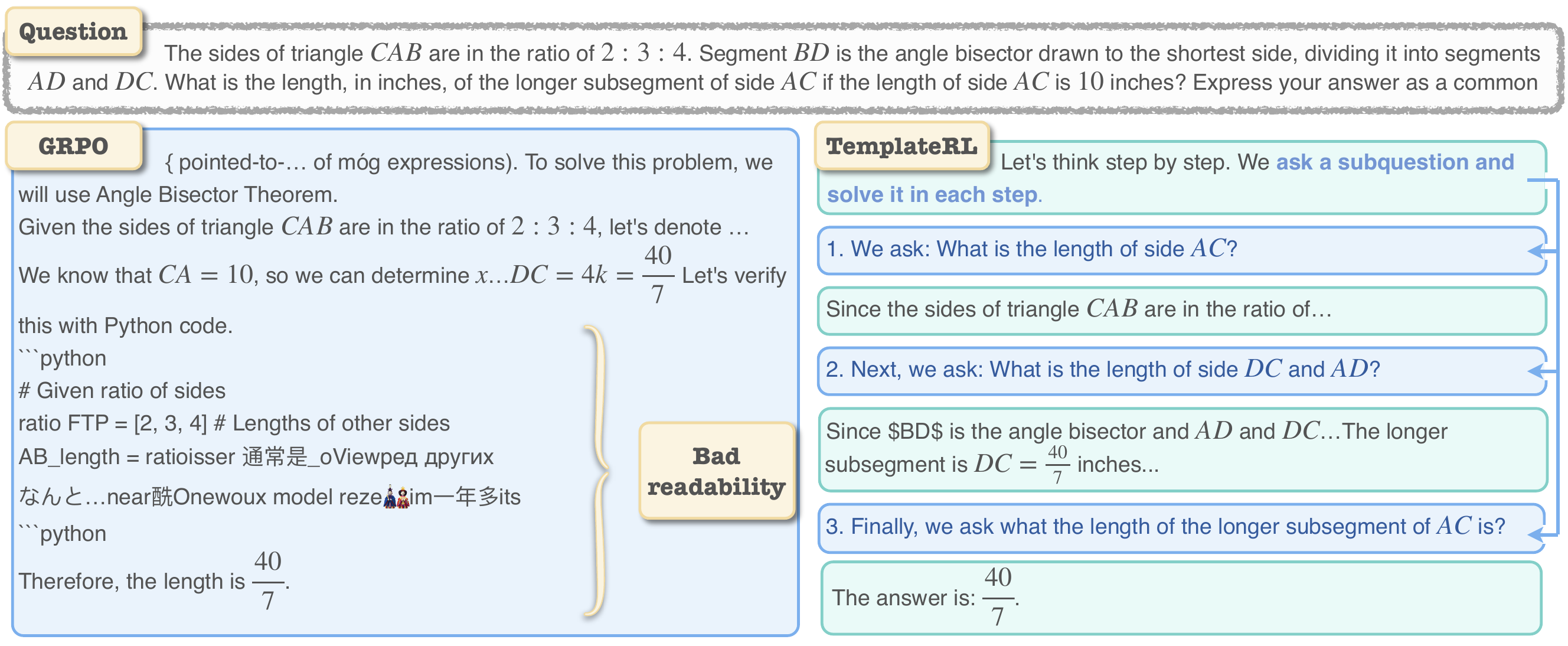}}
\caption{\textbf{Case Study.} TemplateRL produces a more structured and interpretable reasoning chain with clear steps.}
\label{Case}
\end{center}
\vskip -0.25in
\end{figure*}

\subsection{Dynamic Extensibility}\label{Extensibility}
We further explore the dynamic extensibility of the template library, enabling continuous knowledge updates during both training and inference.

\paragraph{Training-time Template Expansion.}
We explore dynamic library expansion during training by continuously incorporating new templates from successful rollouts. After training on 1k, 2k, 3k, and 5k samples, we extract high-level action sequences from correct samples and add them to the library. As shown in Table~\ref{tab:dynamic_training}, dynamic expansion achieves substantial improvements across all benchmarks, with notable gains on AIME24 (+10.2\%). This indicates that our explicit structured library enables continuous updates and reuse of successful exploration experiences, benefiting experience-intensive domains such as medical reasoning where iterative knowledge updates are necessary.

\paragraph{Test-time Template Expansion.}
Beyond training, we explore library expansion during inference. For each test sample, we retrieve five templates to generate multiple paths, apply majority voting for the answer, and immediately abstract patterns from voted outputs into the library before processing the next sample. Table~\ref{tab:dynamic_inference} shows substantial improvements, demonstrating progressive library enrichment during inference using majority-voted pseudo-supervision. This continuous expansion capability helps for test-time scaling scenarios where models leverage accumulated knowledge from earlier predictions to improve subsequent ones.

\begin{table}[t]
\centering
\resizebox{1.0\linewidth}{!}{
\begin{tabular}{lcccc}
\toprule
Method & AIME24~$\uparrow$ & AMC~$\uparrow$ & GPQA-D~$\uparrow$ & Avg.~$\uparrow$ \\
\midrule
\rowcolor{mygray}
\rowcolor{pink!15}
Fixed-5k & 33.3 & 77.5 & 37.9 & 49.6 \\
\midrule
Dynamic-1k & 16.7 & 52.5 & 20.2 & 29.8 \\
Dynamic-2k & 23.3 & 65.0 & 25.8 & 38.0 \\
Dynamic-5k & 36.7 & 80.0 & 38.9 & 51.9 \\
\bottomrule
\end{tabular}
}
\caption{\textbf{Training-time dynamic extensibility.} We compare the fixed library baseline with dynamic expansion that continuously updates from successful rollouts.}
\label{tab:dynamic_training}
\end{table}

\begin{table}[t]
\centering
\resizebox{1.0\linewidth}{!}{
\begin{tabular}{lcccc}
\toprule
Method & AIME24~$\uparrow$ & AMC~$\uparrow$ & GPQA-D~$\uparrow$ & Avg.~$\uparrow$ \\
\midrule
w/o update & 33.3 & 77.5 & 37.9 & 49.6 \\
\rowcolor[RGB]{236,244,252}
w/ update & 36.7 & 77.5 & 40.4 & 51.6 \\
\rowcolor{pink!15}
\textbf{$\bigtriangleup$ $(\uparrow)$} & +10.2\% & +0\% & +6.6\% & +4.1\% \\
\bottomrule
\end{tabular}
}
\caption{\textbf{Test-time dynamic extensibility.} We process test samples sequentially and expand the template library with majority-voting pseudo-supervision signals.}
\label{tab:dynamic_inference}
\vskip -0.1in
\end{table}

\begin{figure}[!ht]
  \centering
  \begin{subfigure}{0.47\linewidth}
    \centering
    \includegraphics[width=\linewidth]{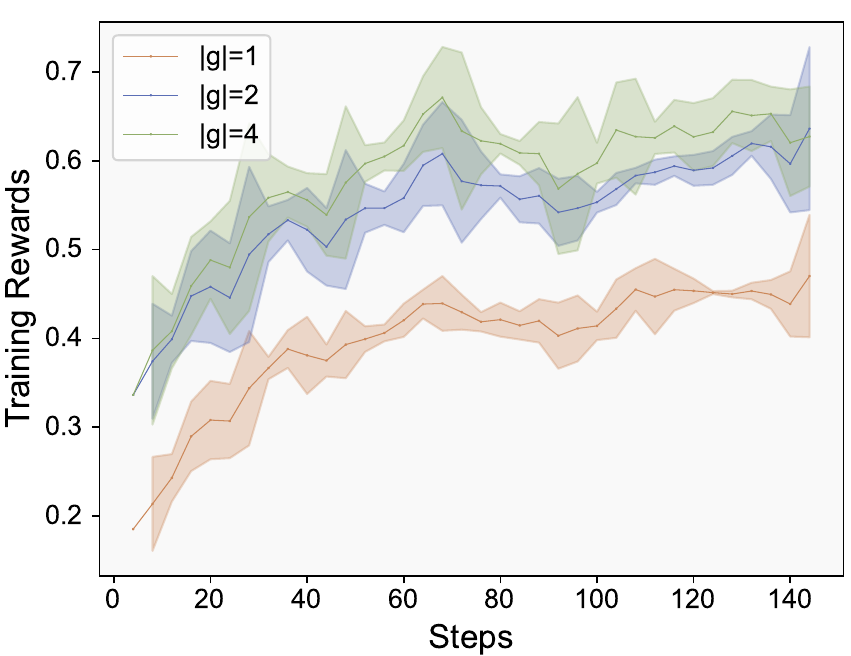}
    \caption{Training Reward Curve}
    \label{fig:subfig_a}
  \end{subfigure}
  \hfill
  \begin{subfigure}{0.45\linewidth}
    \centering
    \includegraphics[width=\linewidth]{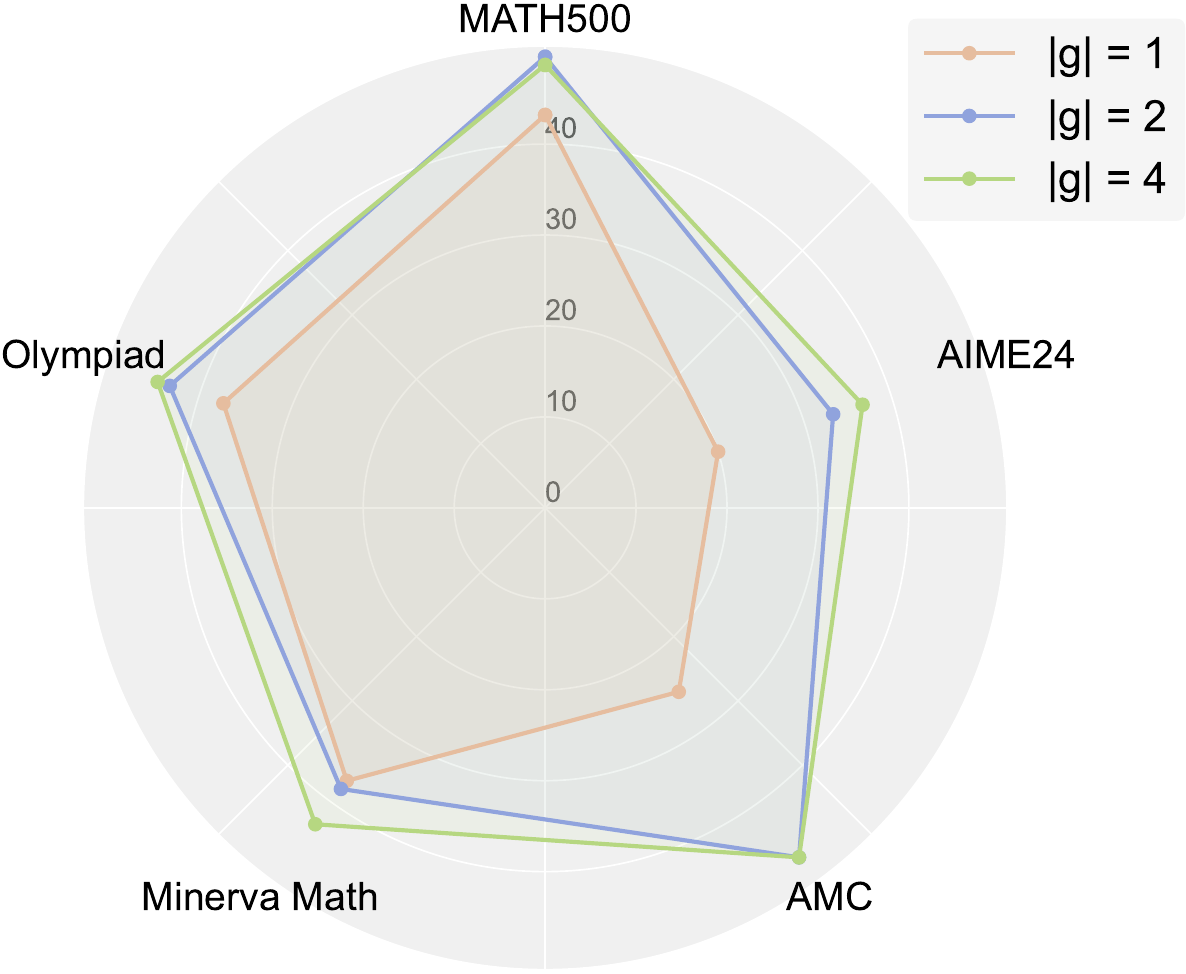}
    \caption{Evaluation Performance}
    \label{fig:subfig_b}
  \end{subfigure}
  \caption{\textbf{Ablation study.} We provide results with different numbers of thought patterns (template guidance).}
  \label{fig:ablation}
  \vskip -0.15in
\end{figure}

\subsection{Ablation Study and Analysis}\label{ablation}
\paragraph{Ablation Study.}
We investigate the effect of template guidance quantity ($|g|=1,2,4$) on training dynamics with fixed total rollouts. Figure~\ref{fig:ablation} shows training rewards and evaluation results across varying templates. For clearer visualization, we uniformly adjust performance on AMC and MATH by subtracting a fixed value without affecting conclusions. We observe that more diverse template guidance often yields higher rewards. We choose \( |g| = 2 \) by default for flexible consideration.

\paragraph{Case Study.}
As shown in Figure~\ref{Case} on a geometry problem, while GRPO produces less readable outputs, TemplateRL first identifies its strategy (e.g., divide-and-conquer), then addresses each subproblem with clear solutions. This demonstrates how template-guided RL training enhances system readability and interpretability.
\paragraph{Other Analysis.}
We provide additional results and analysis in Appendix, including comprehensive OOD results (\ref{D.1}), discussion on template construction strategy (\ref{app:mcts_comparison}), ablation study (\ref{D.8}), statistical analysis (\ref{D.9}), case study (\ref{E}).

\section{Related Work}\label{section5}
\paragraph{RL for LLMs.}
Reinforcement learning (RL) has become a popular paradigm for enhancing LLM reasoning~\citep{guo2025deepseek,team2025kimi,tang2025agent}. Recent research mainly focuses on: (1) algorithmic refinements that address RL's inherent limitations~\citep{yu2025dapo,liu2025understanding}; (2) data-level optimizations~\citep{zuo2025ttrl,wang2025reinforcement}; and (3) exploration enhancement~\citep{wang20258020rulehighentropyminority,cui2025entropy}. However, these methods remain constrained by unstructured self-sampling, lacking mechanisms to leverage structured guidance~\citep{yue2025does}. TemplateRL addresses this by integrating explicit reasoning templates into policy optimization, steering training toward validated strategic patterns.

\vskip -0.2in
\paragraph{Reasoning with Template Guidance.}
Template-based guidance has been explored primarily at inference time, including retrieval-augmented generation~\citep{asai2023retrieval,zhao2024retrieval}, task decomposition~\citep{khot2022decomposed}, and abstract thought patterns~\citep{yang2024buffer,wu2024beyond}. Few works systematically integrate structured templates into RL training. Our approach bridges this gap by constructing a template library and incorporating it during policy optimization. This enables effective learning of both concrete steps and underlying strategic logic, achieving superior performance, flexibility, and transferability.

\section{Conclusion}\label{section6}
We introduce TemplateRL, a template-guided RL framework for LLM reasoning. By constructing structured templates and integrating them during policy optimization, TemplateRL steers training toward validated strategic patterns. Experiments demonstrate significant improvements across models, datasets, and modalities. The framework exhibits cross-domain generalization, dynamic extensibility, and produces interpretable decision chains.

\section*{Limitations}
In this paper, we primarily explore TemplateRL on text and multimodal reasoning tasks. In future work, we plan to extend the framework to more practical domains such as robotic control, scientific discovery, and embodied AI agents. Applying structured template guidance to real-world scenarios is expected to yield broader impact and demonstrate the generalizability of structured guidance across physical and interactive environments.

\section*{Acknowledgments}
This work is supported by the National Natural Science Foundation of China (No. U2436210, No. 62322120).

\bibliography{custom}

@article{tang2025agent,
  title={Agent kb: Leveraging cross-domain experience for agentic problem solving},
  author={Tang, Xiangru and Qin, Tianrui and Peng, Tianhao and Zhou, Ziyang and Shao, Daniel and Du, Tingting and Wei, Xinming and Xia, Peng and Wu, Fang and Zhu, He and others},
  journal={arXiv preprint arXiv:2507.06229},
  year={2025}
}

@article{chen-etal-2024-retrieval-style,
    title = "Retrieval-style In-context Learning for Few-shot Hierarchical Text Classification",
    author = "Chen, Huiyao  and
      Zhao, Yu  and
      Chen, Zulong  and
      Wang, Mengjia  and
      Li, Liangyue  and
      Zhang, Meishan  and
      Zhang, Min",
    journal = "Transactions of the Association for Computational Linguistics",
    volume = "12",
    year = "2024",
    address = "Cambridge, MA",
    publisher = "MIT Press",
    pages = "1214--1231"
}

@article{miao2025blindguard,
  title={Blindguard: Safeguarding llm-based multi-agent systems under unknown attacks},
  author={Miao, Rui and Liu, Yixin and Wang, Yili and Shen, Xu and Tan, Yue and Dai, Yiwei and Pan, Shirui and Wang, Xin},
  journal={arXiv preprint arXiv:2508.08127},
  year={2025}
}

@inproceedings{shen2025understanding,
  title={Understanding the information propagation effects of communication topologies in llm-based multi-agent systems},
  author={Shen, Xu and Liu, Yixin and Dai, Yiwei and Wang, Yili and Miao, Rui and Tan, Yue and Pan, Shirui and Wang, Xin},
  booktitle={Proceedings of the 2025 Conference on Empirical Methods in Natural Language Processing},
  pages={12358--12372},
  year={2025}
}

@inproceedings{wu-etal-2025-pandoras,
    title = "Pandora{'}s Box or Aladdin{'}s Lamp: A Comprehensive Analysis Revealing the Role of {RAG} Noise in Large Language Models",
    author = "Wu, Jinyang  and
      Zhang, Shuai  and
      Che, Feihu  and
      Feng, Mingkuan  and
      Shao, Pengpeng  and
      Tao, Jianhua",
    booktitle = "Proceedings of the 63rd Annual Meeting of the Association for Computational Linguistics (Volume 1: Long Papers)",
    month = jul,
    year = "2025",
    address = "Vienna, Austria",
    publisher = "Association for Computational Linguistics",
    pages = "5019--5039",
}

@article{chen2025beyond,
  title={Beyond Chunking: Discourse-Aware Hierarchical Retrieval for Long Document Question Answering},
  author={Chen, Huiyao and Yang, Yi and Li, Yinghui and Zhang, Meishan and Hu, Baotian and Zhang, Min},
  journal={arXiv preprint arXiv:2506.06313},
  year={2025}
}

@article{liao2025moa,
  title={MOA: Multi-Objective Alignment for Role-Playing Agents},
  author={Liao, Chonghua and Wang, Ke and Wu, Yuchuan and Huang, Fei and Li, Yongbin},
  journal={arXiv preprint arXiv:2512.09756},
  year={2025}
}

@article{kang2025entropy,
  title={Entropy regularizing activation: Boosting continuous control, large language models, and image classification with activation as entropy constraints},
  author={Kang, Zilin and Liao, Chonghua and Xu, Tingqiang and Xu, Huazhe},
  journal={arXiv preprint arXiv:2510.08549},
  year={2025}
}

@article{feng2025two,
  title={Two-Stage Regularization-Based Structured Pruning for LLMs},
  author={Feng, Mingkuan and Wu, Jinyang and Liu, Siyuan and Zhang, Shuai and Jin, Ruihan and Che, Feihu and Shao, Pengpeng and Wen, Zhengqi and Tao, Jianhua},
  journal={arXiv preprint arXiv:2505.18232},
  year={2025}
}

@article{wu2026atlas,
  title={Atlas: Orchestrating Heterogeneous Models and Tools for Multi-Domain Complex Reasoning},
  author={Wu, Jinyang and Zhai, Guocheng and Jin, Ruihan and Yuan, Jiahao and Shen, Yuhao and Zhang, Shuai and Wen, Zhengqi and Tao, Jianhua},
  journal={arXiv preprint arXiv:2601.03872},
  year={2026}
}

@article{shen2026double,
  title={Double: Breaking the Acceleration Limit via Double Retrieval Speculative Parallelism},
  author={Shen, Yuhao and Liu, Tianyu and Shen, Junyi and Wu, Jinyang and Kong, Quan and Huan, Li and Wang, Cong},
  journal={arXiv preprint arXiv:2601.05524},
  year={2026}
}

@article{wu2026spark,
  title={Spark: Strategic Policy-Aware Exploration via Dynamic Branching for Long-Horizon Agentic Learning},
  author={Wu, Jinyang and Yang, Shuo and Yang, Changpeng and Shen, Yuhao and Zhang, Shuai and Wen, Zhengqi and Tao, Jianhua},
  journal={arXiv preprint arXiv:2601.20209},
  year={2026}
}

@article{xu2026odysseyarena,
  title={OdysseyArena: Benchmarking Large Language Models For Long-Horizon, Active and Inductive Interactions},
  author={Xu, Fangzhi and Yan, Hang and Sun, Qiushi and Wu, Jinyang and Huang, Zixian and Huang, Muye and Gong, Jingyang and Ding, Zichen and Cheng, Kanzhi and Wang, Yian and others},
  journal={arXiv preprint arXiv:2602.05843},
  year={2026}
}

@inproceedings{
wan2025rema,
title={Re{MA}: Learning to Meta-Think for {LLM}s with Multi-agent Reinforcement Learning},
author={Ziyu Wan and Yunxiang LI and Xiaoyu Wen and Yan Song and Hanjing Wang and Linyi Yang and Mark Schmidt and Jun Wang and Weinan Zhang and Shuyue Hu and Ying Wen},
booktitle={The Thirty-ninth Annual Conference on Neural Information Processing Systems},
year={2025}
}

@article{wang20258020rulehighentropyminority,
      title={Beyond the 80/20 Rule: High-Entropy Minority Tokens Drive Effective Reinforcement Learning for LLM Reasoning}, 
      author={Shenzhi Wang and Le Yu and Chang Gao and Chujie Zheng and others},
      year={2025},
      journal={arXiv preprint arXiv:2506.01939},
}

@article{cui2025entropy,
  title={The Entropy Mechanism of Reinforcement Learning for Reasoning Language Models},
  author={Cui, Ganqu and Zhang, Yuchen and Chen, Jiacheng and Yuan, Lifan and Wang, Zhi and Zuo, Yuxin and Li, Haozhan and Fan, Yuchen and Chen, Huayu and Chen, Weize and others},
  journal={arXiv preprint arXiv:2505.22617},
  year={2025}
}

@article{guo2025deepseek,
  title={Deepseek-r1: Incentivizing reasoning capability in llms via reinforcement learning},
  author={Guo, Daya and Yang, Dejian and Zhang, Haowei and Song, Junxiao and Zhang, Ruoyu and Xu, Runxin and Zhu, Qihao and Ma, Shirong and Wang, Peiyi and Bi, Xiao and others},
  journal={arXiv preprint arXiv:2501.12948},
  year={2025}
}

@article{shao2024deepseekmath,
  title={Deepseekmath: Pushing the limits of mathematical reasoning in open language models},
  author={Shao, Zhihong and Wang, Peiyi and Zhu, Qihao and Xu, Runxin and Song, Junxiao and Bi, Xiao and Zhang, Haowei and Zhang, Mingchuan and Li, YK and Wu, Y and others},
  journal={arXiv preprint arXiv:2402.03300},
  year={2024}
}

@book{sutton1998reinforcement,
  title={Reinforcement learning: An introduction},
  author={Sutton, Richard S and Barto, Andrew G and others},
  volume={1},
  number={1},
  year={1998},
  publisher={MIT press Cambridge}
}

@article{zhang2023evaluating,
  title={Evaluating the performance of large language models on gaokao benchmark},
  author={Zhang, Xiaotian and Li, Chunyang and Zong, Yi and Ying, Zhengyu and He, Liang and Qiu, Xipeng},
  journal={arXiv preprint arXiv:2305.12474},
  year={2023}
}

@inproceedings{gpqa,
      title={{GPQA}: A Graduate-Level Google-Proof Q\&A Benchmark},
      author={David Rein and Betty Li Hou and Asa Cooper Stickland and Jackson Petty and Richard Yuanzhe Pang and Julien Dirani and Julian Michael and Samuel R. Bowman},
      booktitle={First Conference on Language Modeling},
      year={2024},
      url={https://openreview.net/forum?id=Ti67584b98}
}

@article{mmlu_pro,
  title={Mmlu-pro: A more robust and challenging multi-task language understanding benchmark},
  author={Wang, Yubo and Ma, Xueguang and Zhang, Ge and Ni, Yuansheng and Chandra, Abhranil and Guo, Shiguang and Ren, Weiming and Arulraj, Aaran and He, Xuan and Jiang, Ziyan and others},
  journal={arXiv preprint arXiv:2406.01574},
  year={2024}
}

@article{cobbe2021training,
  title={Training verifiers to solve math word problems},
  author={Cobbe, Karl and Kosaraju, Vineet and Bavarian, Mohammad and others},
  journal={arXiv preprint arXiv:2110.14168},
  year={2021}
}

@inproceedings{dataset_olympiad,
  title={OlympiadBench: A Challenging Benchmark for Promoting AGI with Olympiad-Level Bilingual Multimodal Scientific Problems},
  author={He, Chaoqun and Luo, Renjie and Bai, Yuzhuo and Hu, Shengding and Thai, Zhen and Shen, Junhao and Hu, Jinyi and Han, Xu and Huang, Yujie and Zhang, Yuxiang and others},
  booktitle={Proceedings of the 62nd Annual Meeting of the Association for Computational Linguistics (Volume 1: Long Papers)},
  pages={3828--3850},
  year={2024}
}

@article{dataset_minerva,
  title={Solving quantitative reasoning problems with language models},
  author={Lewkowycz, Aitor and Andreassen, Anders and Dohan, David and Dyer, Ethan and Michalewski, Henryk and Ramasesh, Vinay and Slone, Ambrose and Anil, Cem and Schlag, Imanol and Gutman-Solo, Theo and others},
  journal={Advances in Neural Information Processing Systems},
  volume={35},
  pages={3843--3857},
  year={2022}
}

@misc{li2024numinamath,
  author       = {Jia Li and Edward Beeching and Lewis Tunstall and Ben Lipkin and Roman Soletskyi and Shengyi Huang and Kashif Rasul and Longhui Yu and Albert Q. Jiang and Ziju Shen and others},
  title        = {Numinamath: The largest public dataset in AI4Maths with 860k pairs of competition math problems and solutions},
  year         = {2024},
  howpublished = {\url{https://huggingface.co/datasets/Numinamath}}
}

@article{zeng2025simplerl,
  title={Simplerl-zoo: Investigating and taming zero reinforcement learning for open base models in the wild},
  author={Zeng, Weihao and Huang, Yuzhen and Liu, Qian and Liu, Wei and He, Keqing and Ma, Zejun and He, Junxian},
  journal={arXiv preprint arXiv:2503.18892},
  year={2025}
}

@inproceedings{
hendrycks2021measuring,
title={Measuring Mathematical Problem Solving With the {MATH} Dataset},
author={Dan Hendrycks and Collin Burns and Saurav Kadavath and Akul Arora and Steven Basart and Eric Tang and Dawn Song and Jacob Steinhardt},
booktitle={Thirty-fifth Conference on Neural Information Processing Systems Datasets and Benchmarks Track (Round 2)},
year={2021}
}

@article{qwen2.5_math,
  title={Qwen2. 5-math technical report: Toward mathematical expert model via self-improvement},
  author={Yang, An and Zhang, Beichen and Hui, Binyuan and others},
  journal={arXiv preprint arXiv:2409.12122},
  year={2024}
}

@article{yu2025dapo,
  title={Dapo: An open-source llm reinforcement learning system at scale},
  author={Yu, Qiying and Zhang, Zheng and Zhu, Ruofei and Yuan, Yufeng and Zuo, Xiaochen and Yue, Yu and Fan, Tiantian and Liu, Gaohong and Liu, Lingjun and Liu, Xin and others},
  journal={arXiv preprint arXiv:2503.14476},
  year={2025}
}

@article{liu2025understanding,
  title={Understanding r1-zero-like training: A critical perspective},
  author={Liu, Zichen and Chen, Changyu and Li, Wenjun and Qi, Penghui and Pang, Tianyu and Du, Chao and Lee, Wee Sun and Lin, Min},
  journal={arXiv preprint arXiv:2503.20783},
  year={2025}
}

@Book{Kahneman2011,
  author    = {Daniel Kahneman},
  title     = {Thinking, Fast and Slow},
  publisher = {Farrar, Straus and Giroux},
  year      = {2011},
  address   = {New York, NY},
  isbn      = {978-0374275631}
}

@article{jaech2024openai,
  title={Openai o1 system card},
  author={Jaech, Aaron and Kalai, Adam and Lerer, Adam and Richardson, Adam and El-Kishky, Ahmed and Low, Aiden and Helyar, Alec and Madry, Aleksander and Beutel, Alex and Carney, Alex and others},
  journal={arXiv preprint arXiv:2412.16720},
  year={2024}
}

@article{team2025kimi,
  title={Kimi k1. 5: Scaling reinforcement learning with llms},
  author={Team, Kimi and Du, Angang and Gao, Bofei and Xing, Bowei and Jiang, Changjiu and Chen, Cheng and Li, Cheng and Xiao, Chenjun and Du, Chenzhuang and Liao, Chonghua and others},
  journal={arXiv preprint arXiv:2501.12599},
  year={2025}
}

@article{hu2025open,
  title={Open-reasoner-zero: An open source approach to scaling up reinforcement learning on the base model},
  author={Hu, Jingcheng and Zhang, Yinmin and Han, Qi and Jiang, Daxin and Zhang, Xiangyu and Shum, Heung-Yeung},
  journal={arXiv preprint arXiv:2503.24290},
  year={2025}
}

@article{yue2025does,
  title={Does Reinforcement Learning Really Incentivize Reasoning Capacity in LLMs Beyond the Base Model?},
  author={Yue, Yang and Chen, Zhiqi and Lu, Rui and Zhao, Andrew and Wang, Zhaokai and Song, Shiji and Huang, Gao},
  journal={arXiv preprint arXiv:2504.13837},
  year={2025}
}

@article{wang2025reinforcement,
  title={Reinforcement Learning for Reasoning in Large Language Models with One Training Example},
  author={Wang, Yiping and Yang, Qing and Zeng, Zhiyuan and others},
  journal={arXiv preprint arXiv:2504.20571},
  year={2025}
}

@article{achiam2023gpt,
  title={Gpt-4 technical report},
  author={Achiam, Josh and Adler, Steven and Agarwal, Sandhini and Ahmad, Lama and Akkaya, Ilge and Aleman, Florencia Leoni and Almeida, Diogo and Altenschmidt, Janko and Altman, Sam and Anadkat, Shyamal and others},
  journal={arXiv preprint arXiv:2303.08774},
  year={2023}
}

@article{grattafiori2024llama,
  title={The llama 3 herd of models},
  author={Grattafiori, Aaron and Dubey, Abhimanyu and Jauhri, Abhinav and Pandey, Abhinav and Kadian, Abhishek and Al-Dahle, Ahmad and Letman, Aiesha and Mathur, Akhil and Schelten, Alan and Vaughan, Alex and others},
  journal={arXiv preprint arXiv:2407.21783},
  year={2024}
}

@article{zuo2025ttrl,
  title={TTRL: Test-Time Reinforcement Learning},
  author={Zuo, Yuxin and Zhang, Kaiyan and Qu, Shang and Sheng, Li and Zhu, Xuekai and Qi, Biqing and Sun, Youbang and Cui, Ganqu and Ding, Ning and Zhou, Bowen},
  journal={arXiv preprint arXiv:2504.16084},
  year={2025}
}

@inproceedings{
paglieri2025balrog,
title={{BALROG}: Benchmarking Agentic {LLM} and {VLM} Reasoning On Games},
author={Davide Paglieri and Bart{\l}omiej Cupia{\l} and others},
booktitle={The Thirteenth International Conference on Learning Representations},
year={2025},
url={https://openreview.net/forum?id=fp6t3F669F}
}

@Article{Jaffe23,
  author = 	 "P. I. Jaffe and R. A. Poldrack and R. J. Schafer and others.",
  title = 	 "Modelling human behaviour in cognitive tasks with latent dynamical systems",
  journal =	 "Nature Human Behaviour",
  year =	 "2023",
  volume =	 "7",
  number =	 "",
  pages =	 "986--1000"
}

@article{da2023system,
  title={System 1 vs. System 2 Thinking},
  author={Da Silva, Sergio},
  journal={Psych},
  volume={5},
  number={4},
  pages={1057--1076},
  year={2023},
  publisher={MDPI}
}

@article{qin2024o1,
  title={O1 Replication Journey: A Strategic Progress Report--Part 1},
  author={Qin, Yiwei and Li, Xuefeng and Zou, Haoyang and Liu, Yixiu and Xia, Shijie and Huang, Zhen and Ye, Yixin and Yuan, Weizhe and Liu, Hector and Li, Yuanzhi and others},
  journal={arXiv preprint arXiv:2410.18982},
  year={2024}
}

@InProceedings{11871842_29,
author="Kocsis, Levente
and Szepesv{\'a}ri, Csaba",
editor="F{\"u}rnkranz, Johannes
and Scheffer, Tobias
and Spiliopoulou, Myra",
title="Bandit Based Monte-Carlo Planning",
booktitle="Machine Learning: ECML 2006",
year="2006",
publisher="Springer Berlin Heidelberg",
address="Berlin, Heidelberg",
pages="282--293",
isbn="978-3-540-46056-5"
}

@inproceedings{
wang2023selfconsistency,
title={Self-Consistency Improves Chain of Thought Reasoning in Language Models},
author={Xuezhi Wang and Jason Wei and Dale Schuurmans and Quoc V Le and Ed H. Chi and Sharan Narang and Aakanksha Chowdhery and Denny Zhou},
booktitle={The Eleventh International Conference on Learning Representations },
year={2023}
}

@inproceedings{NEURIPS2021_d5eca8dc,
 author = {Ye, Weirui and Liu, Shaohuai and Kurutach, Thanard and Abbeel, Pieter and Gao, Yang},
 booktitle = {Advances in Neural Information Processing Systems},
 editor = {M. Ranzato and A. Beygelzimer and Y. Dauphin and P.S. Liang and J. Wortman Vaughan},
 pages = {25476--25488},
 publisher = {Curran Associates, Inc.},
 title = {Mastering Atari Games with Limited Data},
 volume = {34},
 year = {2021}
}

@inproceedings{
zhou2024language,
title={Language Agent Tree Search Unifies Reasoning, Acting, and Planning in Language Models},
author={Andy Zhou and Kai Yan and Michal Shlapentokh-Rothman and Haohan Wang and Yu-Xiong Wang},
booktitle={Forty-first International Conference on Machine Learning},
year={2024}
}

@book{dong2020deep,
  title={Deep Reinforcement Learning: Fundamentals, Research and Applications},
  author={Dong, Hao and Ding, Zihan and Zhang, Shanghang},
  isbn={9789811540950},
  publisher={Springer Singapore},
  year={2020},
  series={eBook Packages: Mathematics and Statistics},
  edition={1},
  volume={1},
  pages={514},
}

@inproceedings{chaslot2008monte,
  title={Monte-carlo tree search: A new framework for game ai},
  author={Chaslot, Guillaume and Bakkes, Sander and Szita, Istvan and Spronck, Pieter},
  booktitle={Proceedings of the AAAI Conference on Artificial Intelligence and Interactive Digital Entertainment},
  volume={4},
  number={1},
  pages={216--217},
  year={2008}
}

@inproceedings{yue2023mmmu,
  title={MMMU: A Massive Multi-discipline Multimodal Understanding and Reasoning Benchmark for Expert AGI},
  author={Xiang Yue and Yuansheng Ni and Kai Zhang and Tianyu Zheng and Ruoqi Liu and Ge Zhang and Samuel Stevens and others},
  booktitle={Proceedings of CVPR},
  year={2024},
}

@book{wilcoxon1992individual,
editor = "Kotz, Samuel and Johnson, Norman L.",
title = "Breakthroughs in Statistics: Methodology and Distribution",
year = 1992,
address = "New York, NY",
publisher = "Springer New York",
}

@InProceedings{10.1007/978-3-031-73337-6_9,
author="Fu, Xingyu
and Hu, Yushi
and Li, Bangzheng
and Feng, Yu
and Wang, Haoyu
and Lin, Xudong
and Roth, Dan
and Smith, Noah A.
and Ma, Wei-Chiu
and Krishna, Ranjay",
editor="Leonardis, Ale{\v{s}}
and Ricci, Elisa
and Roth, Stefan
and Russakovsky, Olga
and Sattler, Torsten
and Varol, G{\"u}l",
title="BLINK: Multimodal Large Language Models Can See but Not Perceive",
booktitle="Computer Vision -- ECCV 2024",
year="2025",
publisher="Springer Nature Switzerland",
address="Cham",
pages="148--166",
isbn="978-3-031-73337-6"
}

@inproceedings{zhang2025mathverse,
  title={Mathverse: Does your multi-modal llm truly see the diagrams in visual math problems?},
  author={Zhang, Renrui and Jiang, Dongzhi and Zhang, Yichi and Lin, Haokun and Guo, Ziyu and Qiu, Pengshuo and Zhou, Aojun and Lu, Pan and Chang, Kai-Wei and Qiao, Yu and others},
  booktitle={European Conference on Computer Vision},
  pages={169--186},
  year={2025},
  organization={Springer}
}

@article{lu2023mathvista,
  title={Mathvista: Evaluating mathematical reasoning of foundation models in visual contexts},
  author={Lu, Pan and Bansal, Hritik and Xia, Tony and Liu, Jiacheng and Li, Chunyuan and Hajishirzi, Hannaneh and Cheng, Hao and Chang, Kai-Wei and Galley, Michel and Gao, Jianfeng},
  journal={arXiv preprint arXiv:2310.02255},
  year={2023}
}

@inproceedings{
wang2024measuring,
title={Measuring Multimodal Mathematical Reasoning with {MATH}-Vision Dataset},
author={Ke Wang and Junting Pan and Weikang Shi and Zimu Lu and Houxing Ren and Aojun Zhou and Mingjie Zhan and Hongsheng Li},
booktitle={The Thirty-eight Conference on Neural Information Processing Systems Datasets and Benchmarks Track},
year={2024}
}

@misc{zheng2025easyr1,
  title        = {EasyR1: An Efficient, Scalable, Multi-Modality RL Training Framework},
  author       = {Zheng, Yaowei and Lu, Junting and Wang, Shenzhi and Feng, Zhangchi and Kuang, Dongdong and Xiong, Yuwen},
  howpublished = {\url{https://github.com/hiyouga/EasyR1}},
  year         = {2025}
}

@inproceedings{lu-etal-2021-inter,
    title = "{I}nter-{GPS}: Interpretable Geometry Problem Solving with Formal Language and Symbolic Reasoning",
    author = "Lu, Pan  and
      Gong, Ran  and
      Jiang, Shibiao  and
      Qiu, Liang  and
      Huang, Siyuan  and
      Liang, Xiaodan  and
      Zhu, Song-Chun",
    booktitle = "Proceedings of the 59th Annual Meeting of the Association for Computational Linguistics and the 11th International Joint Conference on Natural Language Processing (Volume 1: Long Papers)",
    month = aug,
    year = "2021",
    address = "Online",
    publisher = "Association for Computational Linguistics",
    pages = "6774--6786",
}

@misc{grpo,
      title={DeepSeekMath: Pushing the Limits of Mathematical Reasoning in Open Language Models}, 
      author={Zhihong Shao and Peiyi Wang and Qihao Zhu and Runxin Xu and Junxiao Song and Xiao Bi and Haowei Zhang and Mingchuan Zhang and Y. K. Li and Y. Wu and Daya Guo},
      year={2024},
      eprint={2402.03300},
      archivePrefix={arXiv},
      primaryClass={cs.CL},
      url={https://arxiv.org/abs/2402.03300}, 
}

@misc{ahmadian2024rloo,
      title={Back to Basics: Revisiting REINFORCE Style Optimization for Learning from Human Feedback in LLMs}, 
      author={Arash Ahmadian and Chris Cremer and Matthias Gallé and Marzieh Fadaee and Julia Kreutzer and Olivier Pietquin and Ahmet Üstün and Sara Hooker},
      year={2024},
      eprint={2402.14740},
      archivePrefix={arXiv},
      primaryClass={cs.LG},
      url={https://arxiv.org/abs/2402.14740}, 
}

@misc{orz,
      title={Open-Reasoner-Zero: An Open Source Approach to Scaling Up Reinforcement Learning on the Base Model}, 
      author={Jingcheng Hu and Yinmin Zhang and Qi Han and Daxin Jiang and Xiangyu Zhang and Heung-Yeung Shum},
      year={2025},
      eprint={2503.24290},
      archivePrefix={arXiv},
      primaryClass={cs.LG},
      url={https://arxiv.org/abs/2503.24290}, 
}

@article{prime,
  title={Process reinforcement through implicit rewards},
  author={Cui, Ganqu and Yuan, Lifan and Wang, Zefan and Wang, Hanbin and Li, Wendi and He, Bingxiang and Fan, Yuchen and Yu, Tianyu and Xu, Qixin and Chen, Weize and others},
  journal={arXiv preprint arXiv:2502.01456},
  year={2025}
}

@misc{qwen25,
    title = {Qwen2.5: A Party of Foundation Models},
    url = {https://qwenlm.github.io/blog/qwen2.5/},
    author = {{Qwen Team}},
    month = {September},
    year = {2024}
}

@article{schulman2017equivalence,
  title={Equivalence between policy gradients and soft q-learning},
  author={Schulman, John and Chen, Xi and Abbeel, Pieter},
  journal={arXiv preprint arXiv:1704.06440},
  year={2017}
}

@article{wu2024beyond,
  title={Beyond examples: High-level automated reasoning paradigm in in-context learning via mcts},
  author={Wu, Jinyang and Feng, Mingkuan and Zhang, Shuai and Che, Feihu and Wen, Zengqi and Tao, Jianhua},
  journal={arXiv preprint arXiv:2411.18478},
  year={2024}
}

@article{yang2024buffer,
  title={Buffer of thoughts: Thought-augmented reasoning with large language models},
  author={Yang, Ling and Yu, Zhaochen and Zhang, Tianjun and Cao, Shiyi and Xu, Minkai and Zhang, Wentao and Gonzalez, Joseph E and Cui, Bin},
  journal={Advances in Neural Information Processing Systems},
  volume={37},
  pages={113519--113544},
  year={2024}
}

@inproceedings{kocsis2006bandit,
  title={Bandit based monte-carlo planning},
  author={Kocsis, Levente and Szepesv{\'a}ri, Csaba},
  booktitle={European conference on machine learning},
  pages={282--293},
  year={2006},
  organization={Springer}
}

@article{guan2025rstar,
  title={rStar-Math: Small LLMs Can Master Math Reasoning with Self-Evolved Deep Thinking},
  author={Guan, Xinyu and Zhang, Li Lyna and Liu, Yifei and Shang, Ning and Sun, Youran and Zhu, Yi and Yang, Fan and Yang, Mao},
  journal={arXiv preprint arXiv:2501.04519},
  year={2025}
}

@article{qi2024mutual,
  title={Mutual reasoning makes smaller llms stronger problem-solvers},
  author={Qi, Zhenting and Ma, Mingyuan and Xu, Jiahang and Zhang, Li Lyna and Yang, Fan and Yang, Mao},
  journal={arXiv preprint arXiv:2408.06195},
  year={2024}
}

@article{russell1991principles,
  title={Principles of metareasoning},
  author={Russell, Stuart and Wefald, Eric},
  journal={Artificial intelligence},
  volume={49},
  number={1-3},
  pages={361--395},
  year={1991},
  publisher={Elsevier}
}

@article{de2024rational,
  title={Rational metareasoning for large language models},
  author={De Sabbata, C Nicol{\`o} and Sumers, Theodore R and Griffiths, Thomas L},
  journal={arXiv preprint arXiv:2410.05563},
  year={2024}
}

@article{lee2000problem,
  title={Problem complexity: A measure of problem difficulty in algebra by using computer},
  author={Lee, Fong-Lok and Heyworth, Rex},
  journal={EDUCATION JOURNAL-HONG KONG-CHINESE UNIVERSITY OF HONG KONG-},
  volume={28},
  number={1},
  pages={85--108},
  year={2000},
  publisher={Citeseer}
}

@article{embretson2008understanding,
  title={Understanding and quantifying cognitive complexity level in mathematical problem solving items},
  author={Embretson, Susan E and Daniel, Robert C},
  journal={Psychology Science},
  volume={50},
  number={3},
  pages={328},
  year={2008}
}

@article{SALADO2014539,
title = {The Concept of Problem Complexity},
journal = {Procedia Computer Science},
volume = {28},
pages = {539-546},
year = {2014},
note = {2014 Conference on Systems Engineering Research},
author = {Alejandro Salado and Roshanak Nilchiani},
}

@inproceedings{asai2023retrieval,
  title={Retrieval-based language models and applications},
  author={Asai, Akari and Min, Sewon and Zhong, Zexuan and Chen, Danqi},
  booktitle={Proceedings of the 61st Annual Meeting of the Association for Computational Linguistics (Volume 6: Tutorial Abstracts)},
  pages={41--46},
  year={2023}
}

@article{zhao2024retrieval,
  title={Retrieval-augmented generation for ai-generated content: A survey},
  author={Zhao, Penghao and Zhang, Hailin and Yu, Qinhan and Wang, Zhengren and Geng, Yunteng and Fu, Fangcheng and Yang, Ling and Zhang, Wentao and Jiang, Jie and Cui, Bin},
  journal={arXiv preprint arXiv:2402.19473},
  year={2024}
}

@article{khot2022decomposed,
  title={Decomposed prompting: A modular approach for solving complex tasks},
  author={Khot, Tushar and Trivedi, Harsh and Finlayson, Matthew and Fu, Yao and Richardson, Kyle and Clark, Peter and Sabharwal, Ashish},
  journal={arXiv preprint arXiv:2210.02406},
  year={2022}
}

@article{tang2024mathscale,
  title={Mathscale: Scaling instruction tuning for mathematical reasoning},
  author={Tang, Zhengyang and Zhang, Xingxing and Wang, Benyou and Wei, Furu},
  journal={arXiv preprint arXiv:2403.02884},
  year={2024}
}

@article{gandhi2025cognitive,
  title={Cognitive behaviors that enable self-improving reasoners, or, four habits of highly effective stars},
  author={Gandhi, Kanishk and Chakravarthy, Ayush and Singh, Anikait and Lile, Nathan and Goodman, Noah D},
  journal={arXiv preprint arXiv:2503.01307},
  year={2025}
}

\clearpage
\newpage
\appendix
\clearpage
\newpage

\setcounter{tocdepth}{-1}
\addtocontents{toc}{\protect\setcounter{tocdepth}{2}}

\section*{Appendix}
\label{sec:appendix}
In this section, we provide a comprehensive elaboration of the TemplateRL algorithm's technical details. We describe the specific implementation of the Monte Carlo Tree Search algorithm, the construction process of our thought template library, and the adaptive retrieval and instantiation mechanisms for thought patterns during reasoning. Additionally, we provide supplementary experiments and case studies to illustrate our points further. The contents are organized as follows:

\tableofcontents

\vspace{1em}
\section{TL;DR: Main Contributions and Takeaways}\label{tldr}
\begin{tcolorbox}[takeawaysbox]
\begin{enumerate}[leftmargin=1em]
    \vskip 0.1in
    \item \textbf{Structure-Guided RL Paradigm}: Augmenting policy optimization with explicit structured templates significantly improves training efficiency, stability, and generalization over unstructured self-sampling.
    \item \textbf{Transferable High-Level Strategic Patterns}: Abstract reasoning templates exhibit remarkable cross-domain and cross-task transferability, enabling knowledge accumulation and flexible expert intervention through interpretable, editable structures.
    \item \textbf{Effective Learning on Weaker Models}: Template-guided training compensates for limited model capacity, achieving stable training where standard RL frequently collapses.
\end{enumerate}
\end{tcolorbox}

Existing RL methods for LLM reasoning rely primarily on unstructured self-sampling to fit scalar rewards, producing inefficient rollouts that fail to capture transferable problem-solving strategies. While recent work has explored algorithmic refinements, data-level optimizations, and exploration enhancement, no prior approach has systematically integrated structured guidance into policy optimization to address the fundamental limitations of trajectory quality, strategy transferability, and interpretability. Our core contributions are:
\begin{itemize}
    \item \textbf{Novel Structure-Guided RL Framework}: We introduce TemplateRL, which constructs an explicit reasoning template library via offline MCTS on minimal seed data, then seamlessly integrates these structured templates as guidance during online RL training. This design steers policy optimization toward validated strategic patterns.    
    \item \textbf{Superior Performance and Stability}: TemplateRL outperforms GRPO by +99\% on AIME, +41\% on AMC, and +17\% on Minerva Math on a 7B backbone, and consistently improves across scales (1.5B-8B) and architectures (Qwen, Llama). It maintains stable training on weaker models (e.g., Llama-3.2-3B) where GRPO typically collapses.
    \item \textbf{Broad Generalization}: Structured templates transfer effectively across diverse domains (BALROG for agents, GPQA-D for science, MMLU-Pro for knowledge) and modalities (text, vision), highlighting the universality of high-level thought patterns.
    \item \textbf{Interpretability and Dynamic Extensibility}: TemplateRL produces structured reasoning trajectories with identifiable strategic patterns, enabling error diagnosis and expert intervention. The template library is interpretable, directly editable, and supports continuous updates during both training and inference.
\end{itemize}

\section{Theoretical Analysis: Multi-Guidance RL and Template Transfer}
\label{app:theory_grpo_seed_transfer}

\subsection{Notation and Assumptions}
We summarize the notation used in the propositions below.
\begin{itemize}[leftmargin=2pt]
  \item $\pi_\theta$: parametric policy, $\pi_{\mathrm{ref}}$: fixed reference policy.
  \item For a single training query $\mathbf{q}$, GRPO collects a total of $G$ trajectories (trials). These $G$ trajectories are partitioned into $m$ groups (small groups), where $m = |g|$ and we assume for simplicity that $G$ is divisible by $m$; denote the per-group size by $g_s := G/m$.
  \item Within group $h\in\{1,\dots,m\}$, trajectories may be dependent (e.g., because they share the same guidance template or retrieved seed); different groups are assumed approximately independent (e.g., they come from different guidance templates or independent retrievals).
  \item For group $h$, let $X_{h,j}$ be the indicator that the $j$-th trajectory in group $h$ is positive (i.e., has $A>0$). Define the group-success event $Y_h := \mathbf{1}\{\exists j\in[1,g_s]: X_{h,j}=1\}$ (i.e., at least one positive in group $h$).
  \item Denote the per-group success probability by $p_h := \Pr(Y_h=1)$. In a symmetric setting we write $p_{\mathrm{grp}}$ for the common per-group success probability.
  \item Other notation (seed database $\mathcal{S}$, retrieval $\mathrm{Retr}_k$, seed-transfer success probability $r$, etc.) follows the earlier appendix.
\end{itemize}

\subsection{Proof of Proposition 3.1}\label{thm:grouped_sampling}

\noindent\textbf{Proposition 3.1.} (Stability and positive-sample guarantee) \textit{Under the grouped sampling setup above, suppose different groups are independent (or weakly dependent so that group-level covariances are negligible). Then:
\begin{enumerate}
  \item The probability that the entire batch of $G$ trajectories contains at least one positive sample equals
  \[
    P_{\ge1}^{\mathrm{batch}} \;=\; 1 - \prod_{h=1}^m (1 - p_h).
  \]
  In particular, if groups are symmetric ($p_h \equiv p_{\mathrm{grp}}$), then
  \[
    P_{\ge1}^{\mathrm{batch}} = 1 - (1 - p_{\mathrm{grp}})^m,
  \]
  which is strictly increasing in $m$ and approaches 1 as $m\to\infty$ for any fixed $p_{\mathrm{grp}}>0$.
  \item Let $g_{h}$ denote the (vector-valued) gradient contribution from group $h$ (e.g., the average of per-trajectory contributions within the group). Assume group contributions are independent with $\mathbb{E}[g_h] = \bar g_{\mathrm{grp}}$ and $\mathrm{Var}[g_h] = \Sigma_{\mathrm{grp}}$ (bounded). Then the variance of the batch-averaged estimator
  \(
    \hat g = \frac{1}{m}\sum_{h=1}^m g_h
  \)
  satisfies
  \[
    \mathrm{Var}[\hat g] \preceq \frac{1}{m}\Sigma_{\mathrm{grp}} + o\!\left(\tfrac{1}{m}\right),
  \]
  i.e. the leading-order variance scales as $\mathcal{O}(1/m)$ (controlled by the number of independent groups), even if within-group samples are dependent.
\end{enumerate}}
\begin{proof}[Proof sketch]
(1) For each group $h$ define the Bernoulli variable $Y_h$ indicating whether the group contains at least one positive trajectory. By definition $\Pr(Y_h=1)=p_h$. If groups are independent, the probability that no group contains a positive sample equals $\prod_{h=1}^m \Pr(Y_h=0) = \prod_{h=1}^m (1-p_h)$, and the complement yields the stated formula. In the symmetric case set $p_h\equiv p_{\mathrm{grp}}$.

Monotonicity in $m$ holds because multiplying an additional factor $(1-p_{\mathrm{grp}})$ decreases the product, increasing its complement; the limit follows as $(1-p_{\mathrm{grp}})^m\to 0$ when $p_{\mathrm{grp}}>0$.

(2) Let $g_h$ denote the aggregated gradient contribution from group $h$, e.g., $g_h = \tfrac{1}{g_s}\sum_{j=1}^{g_s} g_{h,j}$ where $g_{h,j}$ is the per-trajectory contribution. Even if the $g_{h,j}$ within a group are dependent, $g_h$ is a single random vector (with finite second moments). Under independence across groups,
\begin{align*}
    \mathrm{Var}\!\left(\frac{1}{m}\sum_{h=1}^m g_h\right)&= \frac{1}{m^2}\sum_{h=1}^m \mathrm{Var}[g_h]\\
  &= \frac{1}{m}\left(\frac{1}{m}\sum_{h=1}^m \mathrm{Var}[g_h]\right)\\
  &\preceq \frac{1}{m}\Sigma_{\mathrm{grp}},
\end{align*}
where $\Sigma_{\mathrm{grp}}$ is a uniform bound on $\mathrm{Var}[g_h]$. If group covariances are weak (mixing), their contribution is lower-order, giving the $o(1/m)$ term. Thus the dominant variance reduction is governed by the number of independent groups $m$, not the raw total sample size $G$ when within-group dependence is present. Intuitively: dependent samples within a group provide limited additional independent information, so averaging across independent groups is the effective variance reducer.
\end{proof}
\begin{remark}
The theorem clarifies that when data are collected in dependent clusters (groups) the effective sample size is the number of independent clusters \(m\). Therefore, to achieve variance reduction it is valuable to increase the number of independent groups (e.g., by using multiple distinct guidance templates or distinct retrievals) rather than only increasing correlated samples within a single group.
\end{remark}

\subsection{Proof of Proposition 3.2}\label{thm:grouped_template_transfer}
\noindent\textbf{Proposition 3.2.} (Template transfer improves success probability under grouped sampling) \textit{Consider the grouped sampling with $m$ groups, and suppose group $h$ uses PCC retrieval $\mathrm{Retr}_k^{(h)}(\mathbf{q})$ returning $k$ templates whose best trajectories are used as candidate rollouts within that group. Let $T(\mathbf{q})$ be a template produced by MCTS for a seed query $\mathbf{q}$, the per-mini-group template-transfer success probability is:
\[
  r_h = \Pr\Big(\exists\,T\in\mathrm{Retr}_k^{(h)}(\mathbf{q}) \text{ s.t. } A(T(\mathbf{q}), \mathbf{q})>0\Big).
\]
Assume different groups are independent. Then:
\begin{enumerate}
  \item Per-mini-group: the event that mini-group $h$ contains at least one positive trajectory due to template transfer occurs with probability at least $r_h$. If $r_h > p_h^{\mathrm{policy}}$ (the per-group success probability under policy rollouts), template transfer strictly improves per-group success.
  \item Batch-level: the probability that the full batch has at least one positive sample is
  \[
    P_{\ge1}^{\mathrm{template}} \;=\; 1 - \prod_{h=1}^m (1 - r_h),
  \]
  which exceeds the policy-rollout probability $1 - \prod_{h=1}^m (1 - p_h^{\mathrm{policy}})$ whenever $r_h \ge p_h^{\mathrm{policy}}$ for all $h$ and strict for some $h$.
\end{enumerate}}

\label{thm:grouped_template_transfer}

\begin{proof}[Proof sketch]
(1) and (2) follow from the same complement argument as in the proof of Proposition 3.1 (\ref{thm:grouped_sampling}).
\end{proof}

\subsection{Why template transfer increases success probability}

The key reason why template transfer improves the chance of positive samples is that
MCTS produces high-quality trajectories that approximately maximize expected reward
for their original problem. If a new query $\mathbf{q}$ is similar to a seed query $\mathbf{q}'$,
then the action sequence that solves $\mathbf{q}'$ remains near-optimal for $\mathbf{q}$ under mild regularity assumptions.

\begin{lemma}[Transfer success under smoothness]
\label{lem:transfer_smoothness}
Let $T^\star(\mathbf{q}')$ be a template produced by MCTS for a seed query $\mathbf{q}'$ such that $A(T^\star(\mathbf{q}');\mathbf{q}') > 0$. 
Suppose the advantage function $A(T;\mathbf{q})$ is $L$-Lipschitz continuous in $\mathbf{q}$ for all trajectories $\tau$:
\[
  \big| A(T;\mathbf{q}) - A(T;\mathbf{q}') \big| \le L \, d(\mathbf{q},\mathbf{q}')
\]
for some distance metric $d$. Then, if
\[
  A(T^\star(\mathbf{q}');\mathbf{q}') > L \, d_{}(\mathbf{q},\mathbf{q}'),
\]
it follows that $A(T^\star(\mathbf{q}');\mathbf{q}) > 0$.
\end{lemma}

\begin{proof}[Proof sketch]
By Lipschitz continuity:
\[
  A(T^\star(\mathbf{q}');\mathbf{q}) 
  \ge A(T^\star(\mathbf{q}');\mathbf{q}') - L\, d_{}(\mathbf{q},\mathbf{q}').
\]
If the right-hand side is positive, the inequality implies the transferred trajectory still has positive advantage on $\mathbf{q}$.
\end{proof}

\begin{remark}[Template transfer as a high-probability event]
MCTS templates $T^\star(\mathbf{q}')$ are approximately optimal for $\mathbf{q}'$ and often satisfy $A(T^\star(\mathbf{q}');\mathbf{q}') \gg 0$.
Thus, for new queries $\mathbf{q}$ that are close to $\mathbf{q}'$ (small $d_{\text{PCC}}(\mathbf{q},\mathbf{q}')$), the inequality in Lemma~\ref{lem:transfer_smoothness} is likely to hold.
This implies that retrieving high-quality templates from similar seed problems leads to transferred trajectories with positive advantage on the new problem.
Consequently, the per-group template success probability $r_h$ is strictly higher than the baseline policy success probability $p_h^{\mathrm{policy}}$.
\end{remark}

\begin{figure*}[ht!]
\begin{center}
\centerline{\includegraphics[width=0.96\textwidth]{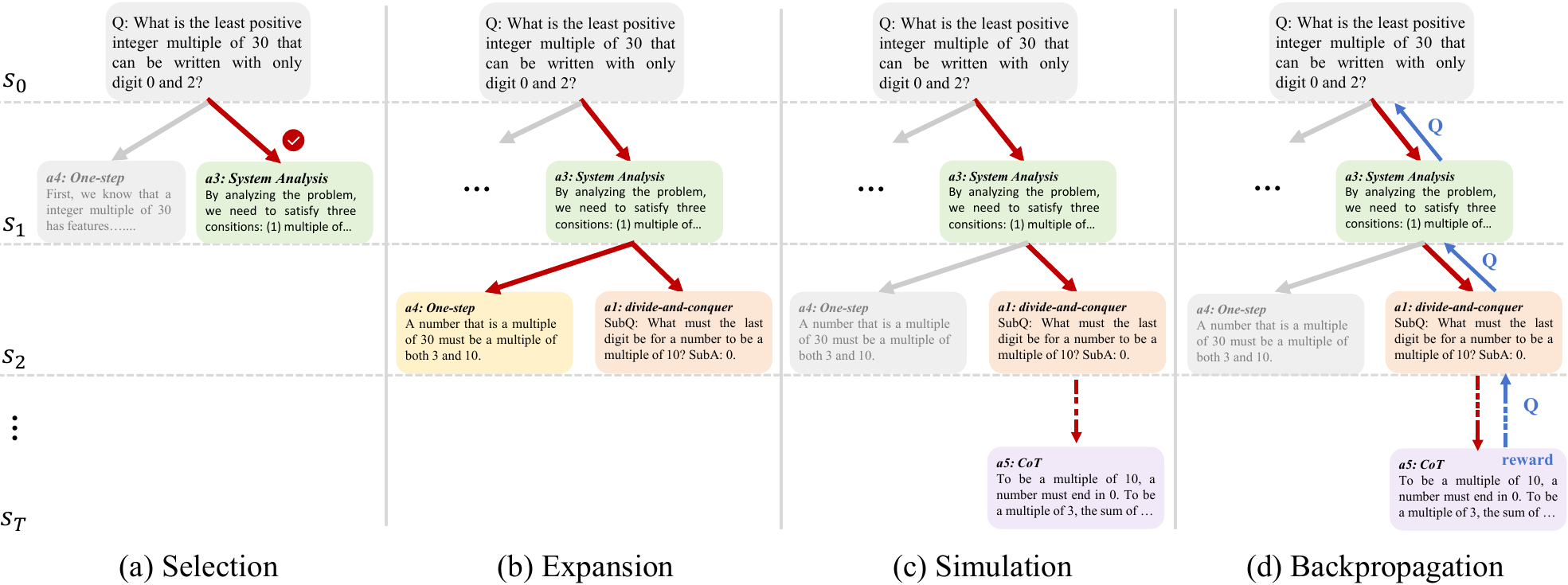}}
\caption{An illustration of four phases in an iteration of MCTS for complex reasoning tasks.}
\label{mcts_examples}
\end{center}
\end{figure*}

\section{More Details about TemplateRL}\label{B}
\subsection{Monte Carlo Tree Search (MCTS)}\label{B.1}
As a heuristic search algorithm, MCTS has demonstrated remarkable success in complex reasoning and decision-making environments~\citep{chaslot2008monte,NEURIPS2021_d5eca8dc,zhou2024language}. The algorithm conceptualizes search spaces as tree structures and has achieved significant breakthroughs across various domains, most notably in game-playing AI such as AlphaGo and AlphaZero~\citep{dong2020deep}. As described in Section 2.3 in the main text, we employ MCTS to generate solution trees based on a small set of 500 seed samples.

To implement MCTS effectively, we first define a predefined action set. Understanding human complex reasoning is crucial for modeling cognitive processes~\citep{Jaffe23}. Existing studies distinguish between two cognitive systems: System 1 and System 2~\citep{Kahneman2011,da2023system}. While ``System 1'' represents fast, intuitive, yet error-prone thinking, ``System 2'' involves slow, deliberative thinking with superior performance. With the emergence of advanced models like OpenAI's o1, developing efficient ``System 2'' approaches to emulate human cognitive processes has gained significant research attention~\citep{qin2024o1,guo2025deepseek}.

Inspired by this and following previous work~\cite{wu2024beyond,qi2024mutual}, we introduce five human-like reasoning actions to bridge the gap between model reasoning and human cognition:
\begin{itemize}
    \item \textbf{Divide and Conquer (DC, $a_{1}$)}: Approaching complex problems by breaking them into manageable sub-problems for easier resolution.
    \item \textbf{Self-Reflection (SR, $a_{2}$)}: Assessing and refining prior solutions during the reasoning process to ensure correctness.
    \item \textbf{System Analysis (SA, $a_{3}$)}: Analyzing the overall structure of the problem and identifying the constraints and conditions before addressing it, thereby clarifying task requirements effectively.
    \item \textbf{One-Step Thought (OST, $a_{4}$)}: Aiming to address a single aspect of the problem through a focused and concise reasoning step.
    \item \textbf{Chain-of-Thought (CoT, $a_{5}$)}: Adopting a sequential reasoning process that builds a series of connected logical steps.
\end{itemize}

Based on the above predefined action set $\mathcal{A}=\left\{a_1,...,a_{|\mathcal{A}|}\right\}$ and model $\pi_\theta$, for each question $\mathbf{s}_i \in \mathcal{S}$, MCTS builds a search tree $\mathcal{T}_i$ where: the root node represents question $\mathbf{s}_i$, each edge denotes an action $a\in \mathcal{A}$, and each child node $n$ contains partial solutions generated by $\pi_\theta$ under the corresponding action. A path from root $\mathbf{s}_i$ to leaf node $\mathbf{n}_{i,j,d_{i,j}}$ forms a solution trajectory:
\[
    \mathbf{t}_{i,j}=\big(\mathbf{s}_i, a_{i,j,1}, \mathbf{n}_{i,j,1}, \dots, a_{i,j,d_{i,j}}, \mathbf{n}_{i,j,d_{i,j}}\big),
\]
Each intermediate node $\mathbf{n}_{i,j,\ell}$ is generated based on the cumulative context of its parent nodes and current action: 
\[
\mathbf{n}_{i,j,\ell}=\pi_\theta([\mathbf{s}_i, a_{i,j,1}, \mathbf{n}_{i,j,1}, \dots,a_{i,j,\ell}]).
\]
Specifically, the MCTS algorithm involves an iterative search process with four key steps: selection, expansion, simulation, and backpropagation.

\textit{(1) Selection.} Starting from the root node, we identify optimal nodes for expansion by traversing the tree level-by-level until reaching a leaf node (maximum depth or final answer). To balance exploration and exploitation, we employ Upper Confidence Bounds applied to Trees (UCT)~\citep{11871842_29}:
\[
    \text{UCT}(\mathbf{n}) = Q(\mathbf{n}) + w\sqrt{\frac{\ln N(\mathbf{n}_p)}{N(\mathbf{n})}}, \label{eq:uct}
\]
where $Q(\mathbf{n})$ is the reward value, $N(\mathbf{n})$ is the visit count, $\mathbf{n}_p$ is the parent node, and $w$ is the exploration weight. The node with the highest UCT value is selected.

\textit{(2) Expansion.} The selected node $\mathbf{n}$ is expanded by sampling $n$ actions from $\pi_\theta$ and generating corresponding reasoning outcomes. These $n$ child nodes are then added to the tree.

\textit{(3) Simulation.} Starting from the selected node, we iteratively sample and expand nodes until reaching a terminal state. To enhance efficiency, we implement an early termination strategy based on self-consistency~\citep{wang2023selfconsistency}: if the model's consistency score exceeds threshold $c$ (i.e., $\text{SC}(\mathbf{n}) > c$), the simulation terminates early.

\textit{(4) Backpropagation.} Upon simulation completion, node information is updated along the simulation path. Visit counts are incremented ($N(\mathbf{n}) \leftarrow N(\mathbf{n}) + 1$), and node value $Q(\mathbf{n})$ is propagated backward to its parent node $\mathbf{n}_p$:
\[
    Q(\mathbf{n}_p) \leftarrow (1-\alpha)Q(\mathbf{n}_p)+\alpha Q(\mathbf{n}), \label{eq:backprop}
\]
where $\alpha$ is a discount factor. For terminal nodes, following~\citet{qi2024mutual}, we use the confidence of self-consistency majority voting as the reward value.

\begin{figure*}
  \centering
  \includegraphics[width=0.95\linewidth]{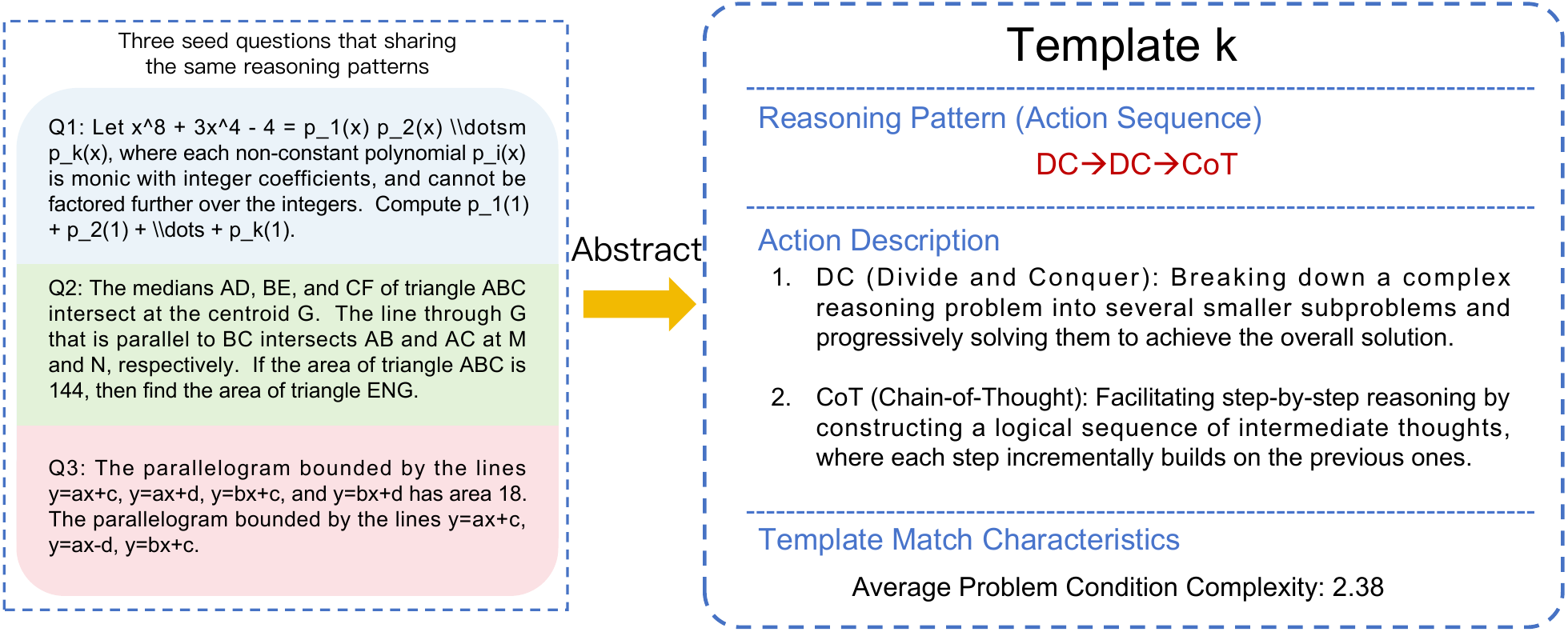}
  \caption{Thought Template Visualization. On the left, the seed dataset contains 3 questions with the same reasoning pattern, ``DC→DC→CoT'', corresponding to template k on the right. The template includes both the reasoning pattern and characteristics used for matching during testing, based on cognitive complexity metrics. When a test question q matches template k, reasoning follows the corresponding pattern ``DC→DC→CoT''.}
  \label{thought_example}
\end{figure*}

Through this process, we obtain diverse solution traces $\mathbb{T}=\left \{ \mathbf{t}_1,\mathbf{t}_2,\dots,\mathbf{t}_t\right \}$ for each question $\mathbf{s}_i \in \mathcal{S}$. MCTS assigns a final reward $R(\mathbf{t}_j|\mathbf{s}_i)$ to each trace $\mathbf{t}_j\in \mathbb{T}$.

Figure \ref{mcts_examples} illustrates the four phases in an iteration, expanding the tree and then updating reward values.

\subsection{Thought Template Library Construction}\label{B.2}

After MCTS exploration, we obtain multiple solution paths for each question $\mathbf{s}_i \in \mathcal{S}$. We then construct a thought template library, which can be used in RL optimization and update during both training and testing process.

\paragraph{Template Construction Process.}
The construction involves five key steps:

\begin{enumerate}
    \item \textbf{Seed Set Selection}:  
    We select a small set of seed problems $\mathcal{S} = \{ s_1, s_2, \dots, s_s \}$ (500 samples in our experiments). These problems are chosen to be complex, while also providing diverse coverage of possible solution strategies.
    
    \item \textbf{MCTS Exploration}:  
    For each seed problem $s_i \in \mathcal{S}$, we apply MCTS to explore the solution space. MCTS recursively simulates different action sequences and evaluates resulting solution paths, generating a solution tree $\mathcal{T}_i$ where each path represents a candidate reasoning trajectory.
    
    \item \textbf{Optimal Path Selection}:  
    From the multiple solution paths generated by MCTS, we select the best trajectory for each question using a balanced scoring metric:
    \[
    \text{Score}(\mathbf{s}_i, \mathbf{t}_j) = b \cdot R(\mathbf{t}_j|\mathbf{s}_i) - (1 - b) \cdot C(\mathbf{t}_j), 
    \]
    where $R(\mathbf{t}_j|\mathbf{s}_i)$ is the reward, $C(\mathbf{t}_j)$ is trajectory complexity (action count), and $b=0.95$ balances solution quality against conciseness. The optimal trace, $\mathbf{t}_{i,\text{best}}$, is selected as the one that maximizes this score: 
    \[
    \mathbf{t}_{i,\text{best}} = \arg\max_{\mathbf{t}_j} \text{Score}(\mathbf{s}_i, \mathbf{t}_j).
    \]
    
    \item \textbf{Template Abstraction}:  
    Since each node in $\mathbf{t}_{i,\text{best}}$ corresponds to an action $a_\ell \in \mathcal{A}$, we extract the action sequence as a high-level template $T_i = (a_1, \dots, a_d)$. For instance, a successful solution might follow the pattern $a_1\rightarrow a_2\rightarrow a_4$ (System Analysis → One-Step Thought → Divide and Conquer). \textbf{Thus, a template is defined as an abstract, high-level sequence of actions that represents a strategic problem-solving approach to a particular task.}

    \item \textbf{Template Aggregation}:  
    To organize the extracted patterns, we use Problem Condition Complexity (PCC)~\citep{lee2000problem,embretson2008understanding} as a categorization metric. PCC quantifies the structural complexity of a problem based on the number of prior conditions and constraints, which can be computed by prompting model $\pi_\theta$ with the question. Templates sharing similar PCC values are grouped together, as problems with comparable structural complexity tend to benefit from similar reasoning strategies. Through this aggregation, each template $\hat{T}_j$ stores both its action pattern and the average PCC of questions sharing this pattern: 
    \[
    \hat{T}_j = (\text{PCC}_{T_j}, T_j),
    \]
    where $\text{PCC}_{T_j} = \frac{1}{|I_j|}\sum_{i \in I_j} \text{PCC}(s_i)$ and $I_j = \{i : T_i = T_j\}$.
\end{enumerate}

\paragraph{Final Template Library.}
The final template library $\mathcal{L} = \{\hat{T}_1, \dots, \hat{T}_m\}$ consists of 103 unique templates derived from 500 seed samples. Each template represents a generalized problem-solving strategy characterized by both its structural pattern (action sequence) and typical problem complexity (average PCC). These templates serve as structured guidance during RL training, enabling the policy to leverage proven strategic patterns when encountering similar problems. The final library thus provides a powerful tool for both training and dynamic adaptation during inference.

For better visualization, we provide a detailed example template in Figure~\ref{thought_example}.

\begin{figure*}[ht!]
% \vskip 0.15in
\begin{center}
\centerline{\includegraphics[width=0.96\textwidth]{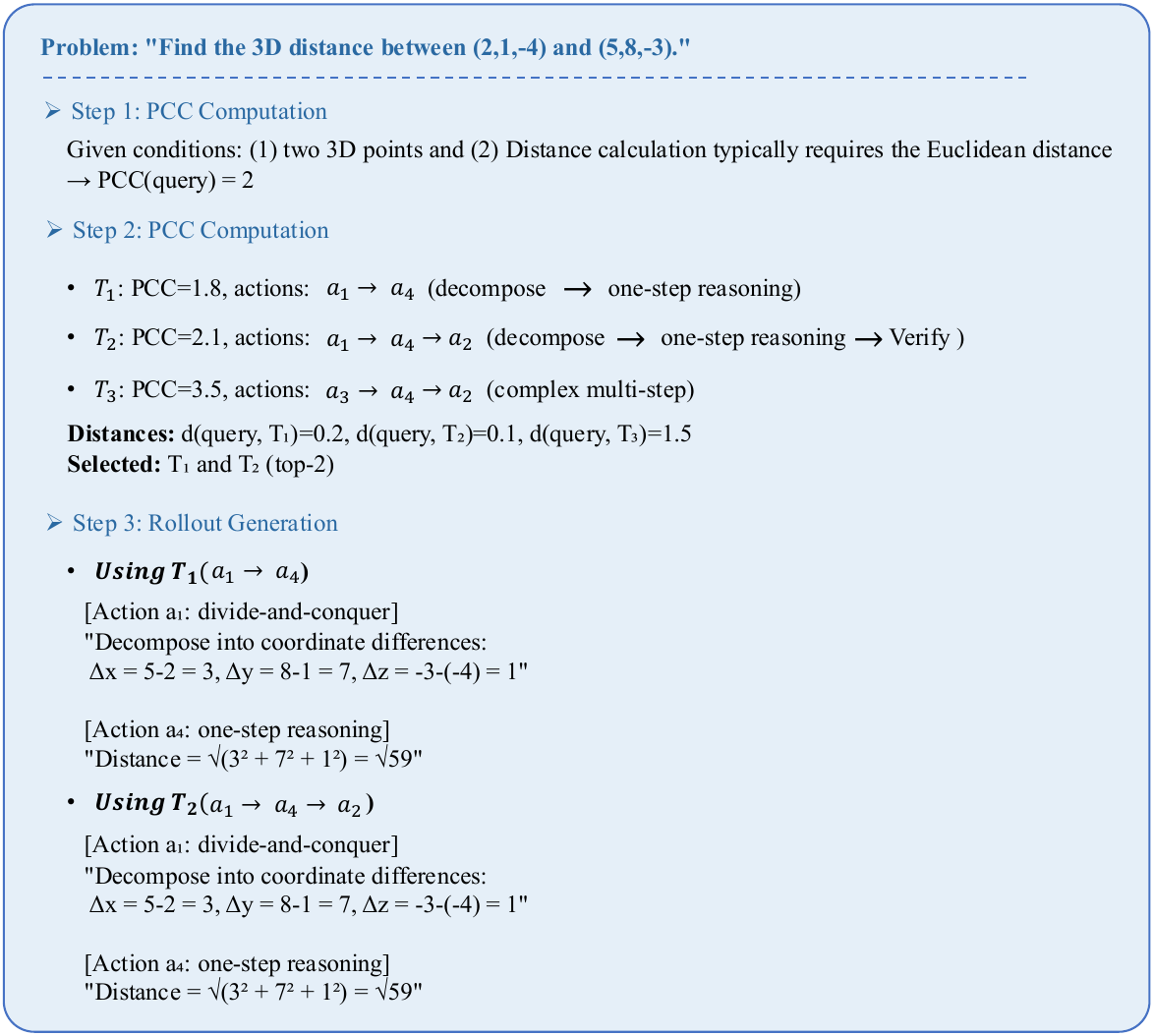}}
\caption{Concrete example of adaptive template retrieval. Given a 3D 
distance calculation problem with $\text{PCC}=2$, the system retrieves 
top-2 templates ($T_1$ and $T_2$) based on complexity similarity while 
filtering out the overly complex $T_3$ ($d=1.5$). The retrieved templates 
guide distinct reasoning trajectories: $T_1$ provides a streamlined solution 
path, while $T_2$ incorporates an additional verification step for enhanced 
reliability.}
\label{concrete_example}
\end{center}
\vskip -0.1in
\end{figure*}

\subsection{Adaptive Template Retrieval}\label{B.3}
During RL training, we employ an adaptive retrieval mechanism to identify 
relevant reasoning strategies from our template library. This approach 
follows meta-reasoning principles~\citep{russell1991principles,de2024rational}, 
which emphasize selecting appropriate strategies based on problem characteristics.

\paragraph{Template Retrieval.}
For each training question $\mathbf{q}_{t}$, we first compute its PCC metric, 
which characterizes the question's structure and complexity. We then calculate 
the distance to each template $\hat{T}_j \in \mathcal{L}$:
\[
d_j = |\text{PCC}_{\mathbf{q}_t} - \text{PCC}_{T_j}|.
\]
This distance quantifies similarity between the current question and problems 
from which each template was derived. Note that, the PCC prompt is: 

\textit{You are an AI assistant to help me rephrase questions by splitting the question context into conditions. In your rephrased question, remember to fully express the information in the original question. That means, you should count the number of distinct prior conditions or given facts that must be satisfied or considered to solve it.}

\textit{Problem: <problem>}

\textit{Number of prior conditions:}

\paragraph{Template Selection and Application.}
We rank templates by distance and select the top-$k$ most similar: 
$\{\hat{T}_{i_1},\dots,\hat{T}_{i_k}\}$. These selected templates contain 
high-level patterns $(a_1,\dots,a_d)$ proven effective for problems with 
similar complexity profiles. During rollout generation, retrieved templates 
serve as structured guidance that balances exploiting known successful 
strategies with model-internal exploration. This adaptive mechanism ensures 
the model leverages appropriate reasoning strategies based on problem 
characteristics, enabling more targeted and effective sampling across diverse 
problem types.

\paragraph{Concrete Example.}
We provide a detailed processing example in Figure~\ref{concrete_example}. This example demonstrates how PCC-based retrieval identifies templates with appropriate complexity levels (rejecting the overly complex $T_3$), and how different templates induce varied reasoning structures. Template $T_1$ provides a concise solution path, while $T_2$ incorporates an additional verification step that enhances answer reliability—both are valid strategies, but $T_2$ may be preferred in high-stakes scenarios where correctness verification is critical. During RL training, the model learns to leverage these structural variations to maximize expected rewards across diverse problem instances.

\begin{figure*}[ht!]
% \vskip 0.15in
\begin{center}
\centerline{\includegraphics[width=0.96\textwidth]{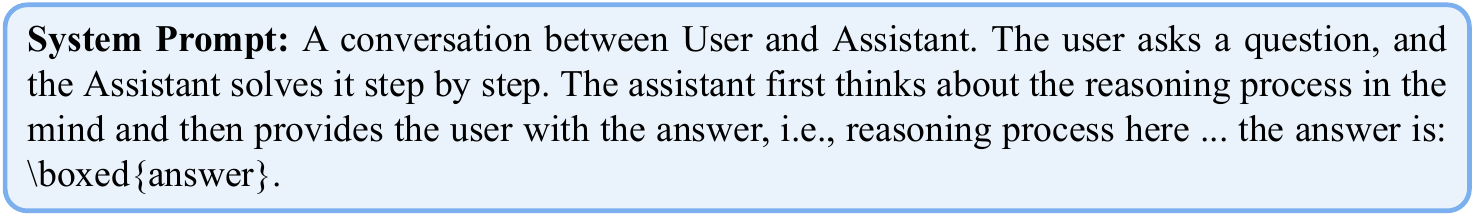}}
\caption{The system prompt used for all experiments.}
\label{system}
\end{center}
\vskip -0.1in
\end{figure*}

\paragraph{Template Updates.}
Regarding template updates, during RL training, we optionally leverage correct rollouts to continually enrich the template library. For each correct rollout, we apply keyword-based or lightweight model-based pattern extraction to automatically identify the underlying action sequence (template), which is then incorporated into the library for continuous expansion. Tables \ref{tab:dynamic_training}-\ref{tab:dynamic_inference} empirically validate the effectiveness of dynamic updates during both training and inference.

\subsection{Discussion on Additional Method Details}\label{app:method_details}
This section further clarifies several key details regarding template 
construction and abstraction.

\paragraph{Template Aggregation via Exact Action Sequence Matching.}
Templates are aggregated through exact matching. Two solution traces share the same template if and only if they follow identical action sequences. For example, all traces following ``$a_1 \rightarrow a_4 
\rightarrow a_3$'' are merged into a single template. The aggregation process treats each unique sequence $T$ as a key, collects all questions matching $T$ with their PCC values, and computes the average PCC to obtain the template's characteristic complexity score, storing the $(T, \text{PCC})$ pair. This exact matching ensures templates represent genuinely identical reasoning structures.

\paragraph{Predefined Actions vs. Automatically Discovered Patterns.}
Our framework employs a two-level design: The action set $\mathcal{A}$ (e.g., divide-and-conquer) comprises predefined atomic reasoning primitives. However, \emph{templates} (sequences of these actions) are automatically discovered through MCTS exploration. For instance, while ``$a_1$'' (divide-and-conquer) is predefined, the template ``$a_1 \rightarrow a_4 \rightarrow a_3$'' emerges from successful solution paths. This design provides interpretability through recognizable reasoning concepts while enabling discovery of effective strategy combinations that may not be obvious to human experts.

\begin{table*}[ht!]
\centering
\resizebox{0.98\linewidth}{!}{
\centering
\begin{tabular}{lccc}
\toprule
\textbf{Model}  & \textbf{GPQA-D$\uparrow$} & \textbf{MMLU-Pro$\uparrow$} & \textbf{BALROG$\uparrow$} \\\midrule
Qwen2.5-Math-7B-Base~\cite{qwen2.5_math}  & 15.2 & 19.2 & 12.5 \\
Qwen2.5-Math-7B-Instruct~\cite{qwen2.5_math} & 28.8 & 36.1 & 15.4\\\midrule
SimpleRL-Zero~\cite{zeng2025simplerl} & 21.2 & 36.9 & 17.4 \\
OpenReasoner-Zero~\cite{orz}     & 30.3 & 54.6 & 28.3 \\
PRIME-Zero~\cite{prime}           & 20.2 & 34.3 & 24.1 \\
Oat-Zero~\cite{liu2025understanding} & 25.8 & 41.8 & 26.2 \\
\midrule
GRPO~\citep{shao2024deepseekmath} & 31.3 & 42.1 & 25.4 \\
\rowcolor{mygray}
TemplateRL (Ours) & \textbf{37.9} & \textbf{49.5} & \textbf{31.5} \\
\rowcolor[RGB]{236,244,252}
\textbf{$\bigtriangleup$ $(\uparrow)$} & +21.1$\%$ & +17.6$\%$ & +24.0$\%$ \\
\bottomrule
\end{tabular}
}
\caption{Detailed results on three representative out-of-distribution benchmarks (Qwen2.5-Math-7B-Base).}
\label{tab:ood}
\end{table*}

\paragraph{Robustness to Paraphrasing through Action-Level Abstraction.}
Template matching operates at the \emph{action level} rather than natural language level. During MCTS, each reasoning step is generated by a specific action prompt and assigned the corresponding action label. Different surface expressions of the same reasoning step (e.g., ``break into x-y coordinates'' vs. ``decompose dimension-wise'' vs. ``handle each axis separately'') are unified through their shared action label $a_1$. When abstracting traces into templates, we extract action sequences rather than literal text, making templates inherently robust to paraphrasing. This contrasts with approaches that directly cluster natural language trajectories, which often require complex similarity metrics to handle linguistic variations.

\section{Experimental Details}\label{C}
In addition to the implementation details in the main paper, we provide supplementary details here:

\paragraph{Baseline Methods.}
For fair comparison, we benchmark TemplateRL with the following baselines on Qwen2.5-Math-7B: (1) \textit{GRPO}~\citep{shao2024deepseekmath}, using the same 5.5k training samples; (2) \textit{SimpleRL-Zero}~\citep{zeng2025simplerl}, applying GRPO to 24k samples from GSM8K and MATH; (3) \textit{OpenReasoner-Zero}~\citep{hu2025open}, employing PPO with 129k samples; (4) \textit{PRIME-Zero}~\citep{prime}, utilizing implicit process rewards on 150k NuminaMath~\citep{li2024numinamath} queries; (5) \textit{Oat-Zero}~\citep{liu2025understanding}, introducing Dr.GRPO to mitigate length bias on 8k MATH questions.

\paragraph{Implementation Details.}
During training, we generate with rollout parameters of temperature=0.8 and top-p=0.95, and a maximum generation of 1500 tokens. The reward function is a binary accuracy metric verified by Math-Verify, defined as \(r(\mathbf{o}) = \mathds{1} \{\text{\(\mathbf{o}\) contains the correct final answer}\}.\) Moreover, we employ a cosine learning rate decay with warm-up. The maximum learning rate is set at \(3\times10^{-6}\), and the warm-up ratio is set at 0.1. We use the same system prompt in Figure~\ref{system} for all RL experiments. All experiments were conducted on a machine running Ubuntu 22.04, equipped with NVIDIA A100-80GB GPUs.

\section{Supplementary Results}\label{D}
\subsection{Out-of-Domain Results}\label{D.1}
We present supplementary out-of-distribution generalization results in Table~\ref{tab:ood}.

\subsection{Extension to More Models}\label{D.2}
In addition to the results presented in the main text based on Qwen2.5-Math-1.5B-Base, Qwen2.5-7B-Instruct, and Llama3.1-8B-Instruct, we further evaluate TemplateRL on two additional models: Qwen2.5-Math-1.5B-Instruct and Llama3.2-3B-Instruct. As illustrated in Table~\ref{table:moremodels_appendix}, TemplateRL achieves an average performance improvement of 3.2\% on Qwen2.5-Math-1.5B-Instruct and 5.6\% on Llama3.2-3B-Instruct.

\begin{table*}[ht!]
\resizebox{0.98\linewidth}{!}{
\centering
\begin{tabular}{lccccccccc}
\toprule
\textbf{Method} & \textbf{AIME24$\uparrow$} & \textbf{AMC$\uparrow$} & \textbf{MATH500$\uparrow$} & \textbf{GSM8K$\uparrow$} & \textbf{Minerva$\uparrow$} & \textbf{Olympiad$\uparrow$} & \textbf{CollegeMath$\uparrow$} & \textbf{Gaokao23$\uparrow$} & \textbf{Avg.$\uparrow$}\\
\midrule
\multicolumn{10}{c}{Qwen2.5-Math-1.5B-Base~\cite{qwen2.5_math}}\\
\midrule
CoT  & 10.0 & 42.5 & 59.0 & 74.6 & 24.3 & 27.6 & 39.5 & 49.6 & 40.9 \\
GRPO   & 13.3 & 40.0 & 66.4 & 74.7  & 25.0 & 30.1  & 40.5 & 52.7 & 42.8 \\
\rowcolor{mygray}
Ours    & \textbf{16.7} & \textbf{55.0} & \textbf{69.0} & \textbf{84.2} & 
                                      \textbf{31.6} &
                                      \textbf{33.6} &  \textbf{47.3} & \textbf{54.8} & \textbf{49.0}\\
\rowcolor[RGB]{236,244,252}
\textbf{$\bigtriangleup$ $(\uparrow)$} & +25.6$\%$ & +37.5$\%$ & +3.9$\%$ & +12.7$\%$ & +26.4$\%$ & +11.7$\%$ & +16.8$\%$ & +4.0$\%$ & +14.5$\%$\\
\midrule
\multicolumn{10}{c}{Qwen2.5-Math-1.5B-Instruct~\cite{qwen2.5_math}}\\
\midrule
CoT  & 6.7 & 47.5 & 68.2 & 76.8  & 28.3 & 36.9  & 47.1 & 63.1 & 46.8 \\
GRPO   & 13.3 & 52.5 & \textbf{76.8} & 85.9 & 28.3 & 36.7 & 45.9 & 65.2 & 50.6 \\
\rowcolor{mygray}
Ours    & \textbf{16.7} & \textbf{55.0} & 76.0 & \textbf{86.5} & \textbf{29.4} &
                                      \textbf{39.7}  &  \textbf{48.3} & \textbf{65.7} & \textbf{52.2} \\
\rowcolor[RGB]{236,244,252}
\textbf{$\bigtriangleup$ $(\uparrow)$} & +25.6$\%$ & +4.8$\%$ & -1.0$\%$ & +1.0$\%$ & +3.9$\%$ & +8.2$\%$ & +5.3$\%$ & +1.0$\%$ & +3.2$\%$\\
\midrule
\multicolumn{10}{c}{Qwen2.5-7B-Instruct~\cite{qwen25}}\\
\midrule
CoT  & 13.3 & 47.5 & 73.2 & 90.0  & 30.5 & 38.8  & 46.9 & 64.2 & 50.5 \\
GRPO  & 13.3 & 57.5 & 76.6 & 90.1 & 32.4 & 36.1 & 44.5 & 62.9 & 51.6 \\
\rowcolor{mygray}
Ours    & \textbf{16.7} & \textbf{67.5} & \textbf{78.0} & \textbf{91.5} & \textbf{36.8} & \textbf{40.6}  &  \textbf{50.6} & \textbf{65.2} & \textbf{55.8} \\
\rowcolor[RGB]{236,244,252}
\textbf{$\bigtriangleup$ $(\uparrow)$} & +25.6$\%$ & +17.4$\%$ & +1.9$\%$ & +1.6$\%$ & +13.6$\%$ & +12.5$\%$ & +13.7$\%$ & +3.7$\%$ & +8.1$\%$\\
\midrule
\multicolumn{10}{c}{Llama-3.1-8B-Instruct~\cite{grattafiori2024llama}}\\
\midrule
CoT  & 3.3 & 20.0 & 36.6 & 77.2  & 16.2 & 15.9 & 13.3 & 29.9 & 26.5\\
GRPO & 3.3 & 22.5 & 45.0 & 82.9 & 21.0 & 16.1 & 31.7 & 40.8 & 32.9 \\
\rowcolor{mygray}
Ours    & \textbf{6.7} & \textbf{30.0} & \textbf{52.2} & \textbf{85.2}  & \textbf{26.8}  & \textbf{17.3} &  \textbf{34.1} & \textbf{42.6} &  \textbf{36.7}\\
\rowcolor[RGB]{236,244,252}
\textbf{$\bigtriangleup$ $(\uparrow)$} & +103.0$\%$ & +33.3$\%$ & +16.0$\%$ & +2.8$\%$ & +27.7$\%$ & +7.5$\%$ & +7.6$\%$ & +4.5$\%$ & +11.6$\%$\\
\midrule
\multicolumn{10}{c}{Llama-3.2-3B-Instruct~\cite{grattafiori2024llama}}\\
\midrule
CoT  & 6.7 & 20.0 & 38.3  & 69.3 & 11.8 & 12.6 & 23.8 & 33.5 & 27.6 \\
GRPO   & 3.3 & 25.0 & 47.8 & 75.2  & 17.6 & 14.5 & 34.1 & 40.8 & 32.2 \\
\rowcolor{mygray}
Ours    & \textbf{6.7} & \textbf{27.5} & \textbf{48.8} & \textbf{78.8}  & \textbf{18.4}  &
                                      \textbf{16.0} &  32.5 &
                                      \textbf{43.1} &
                                      \textbf{34.0}\\
\rowcolor[RGB]{236,244,252}
\textbf{$\bigtriangleup$ $(\uparrow)$} & +103.0$\%$ & +10.0$\%$ & +2.1$\%$ & +4.8$\%$ & +4.6$\%$ & +10.4$\%$ & -4.7$\%$ & +5.7$\%$ & +5.6$\%$\\
\bottomrule
\end{tabular}}
\caption{Accuracy ($\%$) results of different LLMs across eight benchmarks. The best results in each box are highlighted in \textbf{bold}. We provide the relative improvement of our method compared to GRPO.}
\label{table:moremodels_appendix}
\end{table*}

\begin{table*}[ht!]
\resizebox{1.0\linewidth}{!}{
\centering
\begin{tabular}{lccccccccc}
\toprule
\textbf{Method} & \textbf{AIME24$\uparrow$} & \textbf{AMC$\uparrow$} & \textbf{MATH500$\uparrow$} & \textbf{GSM8K$\uparrow$} & \textbf{Minerva$\uparrow$} & \textbf{Olympiad$\uparrow$} & \textbf{CollegeMath$\uparrow$} & \textbf{Gaokao23$\uparrow$} & \textbf{Avg.$\uparrow$}\\
\midrule
\multicolumn{10}{c}{\textit{Qwen2.5-Math-7B-Base}~\cite{qwen2.5_math}}\\
\midrule
RLOO & 30.0 & 50.0 & 73.8 & 82.7 & 35.5 & 36.0 & 39.8 & 64.2 & 51.5 \\
\rowcolor{mygray}
TemplateRL & \textbf{33.3} & \textbf{67.5} & \textbf{78.8} & \textbf{85.8} & \textbf{36.7} & \textbf{41.2} & \textbf{41.8} & \textbf{68.1} & \textbf{56.7} \\
% \rowcolor{mygray}
\rowcolor[RGB]{236,244,252}
\textbf{$\bigtriangleup$ $(\uparrow)$} & +11.0\% & +35.0\% & +6.8\% & +3.7\% & +3.4\% & +14.4\% & +5.0\% & +6.1\% & +10.1\%\\
\midrule
\multicolumn{10}{c}{\textit{Qwen2.5-Math-1.5B-Base}~\cite{qwen2.5_math}}\\
\midrule
RLOO & 20.0 & 50.0 & 68.0 & 82.6 & 28.7 & 32.0 & 41.4 & 51.2 & 46.7 \\
\rowcolor{mygray}
TemplateRL & \textbf{26.7} & \textbf{57.5} & \textbf{75.6} & \textbf{84.7} & \textbf{33.8} & \textbf{33.8} & \textbf{41.8} & \textbf{52.9} & \textbf{50.9} \\
% \rowcolor{mygray}
\rowcolor[RGB]{236,244,252}
\textbf{$\bigtriangleup$ $(\uparrow)$} & +33.5\% & +15.0\% & +11.2\% & +2.5\% & +17.8\% & +5.6\% & +1.0\% & +3.3\% & +9.0\%\\
\bottomrule
\end{tabular}}
\caption{Extension to other RL algorithms. Results demonstrate consistent improvements when integrating thought-augmented training with alternative RL algorithms, highlighting the generalizability of our approach.}
\label{table:rl algorithm}
\end{table*}

\begin{table*}[t]
\centering
\begin{tabular}{l|cc|c}
\toprule
\textbf{Method} & \textbf{AIME25}$\uparrow$ & \textbf{GPQA-Diamond}$\uparrow$ & \textbf{Avg.}$\uparrow$ \\
\midrule
\multicolumn{4}{l}{\textit{Part I: Test-Time Exploration Analysis (w/o RL training)}} \\
\midrule
CoT (pass@16)               & 80.0 & 76.8 & 78.4 \\
\rowcolor{mygray}
Template-Guided Inference (pass@16)   & \textbf{83.3} & \textbf{82.8} & \textbf{83.1} \\
\rowcolor[RGB]{236,244,252}
\quad $\triangle$ (\%)      & +3.3 & +6.0 & +4.7 \\
\midrule
\multicolumn{4}{l}{\textit{Part II: Post-Training Exploration Analysis (after RL)}} \\
\midrule
GRPO (trained, pass@16)     & 83.3 & 80.8 & 82.0 \\
\rowcolor{mygray}
TemplateRL (trained, pass@16) & \textbf{85.7} & \textbf{84.3} & \textbf{85.0} \\
\rowcolor[RGB]{236,244,252}
\quad $\triangle$ vs GRPO (\%) & +2.4 & +4.5 & +3.0 \\
\bottomrule
\end{tabular}
\caption{Exploration Analysis (Pass@16) on highly capable models (Qwen3-32B). Part I examines exploration quality via test-time inference, which mirrors RL rollout sampling. Part II evaluates post-RL performance.}
\label{tab:capable_models}
\end{table*}

\subsection{Extension to Multimodal Domains}\label{D.3}
As shown in Table~\ref{multimodal_results} in the main text, we extend TemplateRL to multimodal domains. We provide more details and analysis here.

\paragraph{Dataset Construction.}
We construct the MMTrain10K dataset based on Geometry3K~\citep{lu-etal-2021-inter}, which contains geometry problems with textual descriptions and visual diagrams. To enhance data diversity and problem complexity distribution, we employ Gemini-2.5-Pro to augment the original dataset by systematically adding or removing auxiliary details from problem statements. This process adjusts problem difficulty while preserving geometric relationships and visual diagrams, ultimately expanding the dataset to 10K samples with varied complexity levels (MMTrain10K).

\paragraph{Implementation Setup.}
We utilize Qwen2.5-VL-3B-Instruct as the base model and follow the EasyR1 pipeline~\citep{zheng2025easyr1} for RL training. Our training configuration maintains consistency with the text-only setting: batch size 128, 16 samples per question, $|g|=2$ structured guidance templates, and maximum generation length of 4096 tokens to accommodate detailed multimodal reasoning processes.

\paragraph{Evaluation Benchmarks.}
We evaluate across five diverse benchmarks: (1) \textbf{Mathematical reasoning}: MathVision~\cite{wang2024measuring}, MathVerse~\citep{zhang2025mathverse}, and MathVista~\citep{lu2023mathvista}; (2) \textbf{General understanding}: MMMU~\citep{yue2023mmmu}; (3) \textbf{Visual perception}: BLINK~\citep{10.1007/978-3-031-73337-6_9}.

\paragraph{Results and Analysis.}
As shown in Table~\ref{multimodal_results}, TemplateRL consistently outperforms both CoT and GRPO across all benchmarks, achieving +8.4\% average improvement over GRPO. Notably, TemplateRL excels on mathematical reasoning tasks: MathVision (+12.0\% vs. GRPO) and MathVerse (+14.8\% vs. GRPO), demonstrating that structured template guidance effectively transfers to multimodal mathematical reasoning. The consistent improvements across diverse tasks (from math to visual perception), validate TemplateRL's core principle: integrating structured reasoning guidance with policy optimization generalizes beyond text modality to multimodal scenarios.

\subsection{Extension to Other RL Algorithms}\label{D.4}
To demonstrate the generalizability of our thought-augmented framework beyond GRPO, we investigate its compatibility with alternative policy optimization algorithms. Specifically, we extend TemplateRL to REINFORCE Leave-One-Out (RLOO)~\citep{ahmadian2024rloo}, a variance-reduced policy gradient method that has shown effectiveness in model fine-tuning. Following our default configuration, we set the number of guidance patterns to 2 (i.e., $|g| = 2$) and maintain consistent experimental settings across both model scales.

As shown in Table~\ref{table:rl algorithm}, TemplateRL consistently enhances RLOO performance across all benchmarks and model configurations. For Qwen2.5-Math-7B-Base, RLOO+TemplateRL achieves substantial improvements over vanilla RLOO, with notable gains on AMC (+17.5 points, +35.0\% relative improvement) and AIME24 (+3.3 points, +11.0\% relative improvement). Similar trends emerge for the smaller Qwen2.5-Math-1.5B-Base model, where RLOO+TemplateRL delivers consistent improvements across all metrics, achieving an average performance boost of +4.2 points (+9.0\% relative improvement). These results demonstrate that the benefits of thought-augmented training extend beyond GRPO to other RL algorithms, highlighting the framework's versatility and broad applicability.

\subsection{Analysis on Highly Capable Models}\label{capable_models}
A natural concern is whether template guidance might restrict the exploration space of larger, more capable models, potentially hindering their inherent reasoning abilities. To address this, we conduct a two-stage analysis on Qwen3-32B with pass@16 evaluation: (1) \textbf{Exploration Quality Analysis}: We compare standard CoT with template-guided test-time inference to assess the quality of 
exploration spaces \emph{before} any RL training-this directly reflects the rollout generation quality during RL sampling; (2) \textbf{Post-Training Performance}: We evaluate models after full RL training using CoT, GRPO, and TemplateRL.

As shown in Table~\ref{tab:capable_models}, template guidance consistently outperforms unrestricted approaches in both stages. In the exploration phase, template-guided sampling achieves 83.3 on AIME25 and 82.8 on GPQA-D, surpassing CoT baseline by 3.3\% and 6.0\%, respectively. This indicates that templates enhance the diversity and quality of candidate solutions. 
After RL training, TemplateRL further amplifies this advantage, demonstrating that structured guidance helps models navigate more efficiently toward high-reward regions without limiting their expressiveness. These results confirm that templates \textbf{enhance rather than restrict} effective exploration of solution spaces, even for larger powerful models. Future work will systematically investigate this scaling behavior across model sizes to further validate these 
preliminary findings.

\subsection{Discussion on Template Construction Strategy}
\label{app:mcts_comparison}
For TemplateRL, a natural question is whether the MCTS-based template construction could be replaced with a simpler end-to-end approach where the policy directly generates abstract templates before producing solutions. Actually, our choice of MCTS is motivated by both theoretical considerations and empirical evidence.

\paragraph{Design Rationale.}
Our MCTS-based approach is grounded in two key principles:

\textit{\underline{(1) Quality of Reasoning Guidance.}}
MCTS-extracted templates are derived from \emph{verified correct reasoning trajectories} obtained through systematic search and validation. The verification process ensures that each template encodes a genuinely effective problem-solving strategy that has been empirically demonstrated to lead to correct solutions. In contrast, end-to-end generation relies on the policy's current capabilities to produce templates, which may:
\begin{itemize}[leftmargin=*,itemsep=0pt]
    \item Generate suboptimal or erroneous templates, especially 
          when the policy is weak or undertrained;
    \item Perpetuate existing biases in the model's reasoning 
          patterns rather than discovering novel strategies;
\end{itemize}

As demonstrated in Figure~\ref{fig:reward_curve}(b) (Section~\ref{training dynamics}), weaker models under standard GRPO training frequently experience training collapse due to sparse positive signals from unstructured exploration. MCTS mitigates this by providing high-quality templates from the outset, 
enabling stable learning even for less capable models.

\textit{\underline{(2) Computational Efficiency.}}
While MCTS appears more complex algorithmically, the computational overhead is minimal in practice. Template construction operates on only 500 seed samples—a 
one-time investment that incurs negligible cost compared to full RL 
training on 5K samples. This small seed set yields a high-quality template library that immediately stabilizes subsequent RL training across thousands of iterations. In contrast, end-to-end approaches must learn template generation \emph{concurrently} with solution generation throughout the entire training process, potentially 
leading to:
\begin{itemize}[leftmargin=*,itemsep=0pt]
    \item Unstable training when template quality is poor;
    \item Wasted computational budgets on low-quality rollouts;
    \item Slower convergence due to dual optimization.
\end{itemize}

\paragraph{Empirical Validation.}
Additionally, we implement and evaluate an end-to-end baseline on Qwen2.5-Math-7B-Base. The end-to-end method follows a two-stage generation process: first, the policy generates an abstract template (e.g., ``divide-and-conquer followed by verification''); second, it produces the complete solution conditioned on this template. Both the template and solution are jointly optimized during RL training. Table~\ref{tab:mcts_vs_end2end_full} presents comprehensive results across three benchmarks. We have two findings:

\begin{table}[h]
\resizebox{1.0\linewidth}{!}{
\centering
% \small
\begin{tabular}{lccc}
\toprule
\textbf{Method} & \textbf{AIME}$\uparrow$ & \textbf{AMC}$\uparrow$ & \textbf{Avg.}$\uparrow$ \\
\midrule
GRPO (baseline)           & 16.7 & 55.0 & 35.9 \\
End-to-End Generation     & 23.3 & 62.5 & 42.9 \\
% \rowcolor{pink!15}
\rowcolor{pink!15}
TemplateRL         & \textbf{33.3} & \textbf{77.5} & \textbf{55.4} \\
\bottomrule
\end{tabular}}
\caption{Comparison of template construction methods on Qwen2.5-Math-7B-Base. End-to-end generates templates via policy before solving; TemplateRL uses MCTS-verified templates.}
\label{tab:mcts_vs_end2end_full}
\end{table}

\begin{table}[t!]
\resizebox{1.0\linewidth}{!}{
\centering
% The column specifier was changed to cccc for four columns
\begin{tabular}{ccccc} 
    \toprule
    Action Space & MATH500~$\uparrow$ & AIME24~$\uparrow$ & AMC~$\uparrow$ & Avg.~$\uparrow$ \\
    \midrule
    $\{a_5\}$ (i.e. CoT) & 76.8 & 23.3 & 67.5 & 55.9\\
    $\{a_1,a_5\}$ & 80.6 & 26.7 & 70.0 & 59.1\\
    $\{a_1,a_4,a_5\}$ & 82.0 & 30.0 & 72.5 & 61.5\\
    $\{a_1,a_3,a_4,a_5\}$ & 82.8 & 30.0 & 77.5 & 63.4\\
    \rowcolor{pink!15}
    $\{a_1,a_2,a_3,a_4,a_5\}$ & 83.4 & 33.3 & 77.5 & 64.7\\
    \bottomrule
\end{tabular}}
\caption{Performance comparison across different action spaces on Qwen2.5-Math-7B-Base. Results demonstrate consistent improvements with expanded spaces.}
\label{action_sensitivity}
\end{table}

\noindent\textbf{(1) MCTS Significantly Outperforms End-to-End.} 
TemplateRL with MCTS-based templates achieves 55.4 average performance, substantially higher than end-to-end generation (42.9). The gap is particularly pronounced on AIME (33.3 vs. 23.3, +43.0\%), which requires complex multi-step reasoning where high-quality template guidance is most critical. This validates our hypothesis that verified template extraction is more effective than direct policy generation.

\noindent\textbf{(2) End-to-End Still Improves Over Vanilla GRPO.} 
The end-to-end method achieves 42.9 vs. GRPO's 35.9 (+19.5\%), suggesting that incorporating structured template generation is beneficial. However, the quality of templates matters significantly: MCTS provides an additional +29.1\% improvement over end-to-end, demonstrating the value of verified guidance.

\paragraph{Future Directions.}
While our results demonstrate MCTS's current superiority, we 
acknowledge the potential of hybrid approaches. Future work could 
explore: (1) Warm-start strategies: Initialize with MCTS-constructed templates to ensure stable early training, then gradually transition to end-to-end refinement once the policy has developed sufficient reasoning capabilities; and (2) Curriculum learning: Start with high-confidence MCTS templates for foundational patterns, then allow the policy to discover novel templates for more complex problems.

\subsection{Computational Overhead Analysis.}
We provide a detailed quantitative breakdown of the additional computational costs 
introduced by TemplateRL relative to standard GRPO.

\textbf{Training overhead.}
\begin{itemize}[leftmargin=*]
    \item \emph{Offline MCTS template construction (one-time cost):} 
    Constructing the template library from 500 seed samples requires approximately 20 minutes on 8$\times$A100-80GB GPUs (0.33 GPU-hours), representing ${<}3\%$ of the total training budget and amortized across all subsequent training runs and model configurations.
    \item \emph{Per-iteration overhead:} 
    PCC calculation adds ${\sim}1$s per batch (batch size 128); structured rollout generation takes ${\sim}1.8\times$ the time of unguided sampling due to iterative action prompting.
    \item \emph{Total training time:} 
    TemplateRL requires ${\sim}18$ GPU-hours (500 steps on 5K samples) vs.\ ${\sim}15$ GPU-hours for GRPO (+20\% overhead). However, this yields a +27.4\% relative performance gain (Table~\ref{tab:main_results}, 55.8 vs.\ 43.8 average accuracy), achieving 3.1 accuracy points per GPU-hour vs.\ 2.9 for GRPO--a favorable efficiency tradeoff.
\end{itemize}

\textbf{Inference overhead.}
\begin{itemize}[leftmargin=*]
    \item \emph{Template retrieval:} ${\sim}0.05$s per question 
    (PCC calculation + distance ranking), negligible compared to generation time.
    \item \emph{Generation time:} 
    Structured rollout takes ${\sim}1.8\times$ unguided generation due to multi-step action prompting, comparable to self-consistency methods that sample multiple paths.
    \item \emph{Dynamic updates (optional):} 
    Add ${\sim}2\%$ additional training time (Table~\ref{tab:dynamic_training}) and ${\sim}0.3$s per question at test-time (Table~\ref{tab:dynamic_inference}), acceptable for accuracy-critical applications.
\end{itemize}

In summary, the +20\% training overhead is well justified by substantial performance gains, and inference overhead remains practical for real-world deployment.

\subsection{Ablation Study on the Action Space}\label{D.7}
To evaluate the sensitivity of our method to the action space used for constructing the thought template library, we conduct a comprehensive ablation study using Qwen2.5-Math-7B-Base as the backbone model. We systematically test various action subsets, ranging from a single action to the complete five-action set, across three representative benchmarks.

As shown in Table~\ref{action_sensitivity}, expanding the action space consistently yields performance improvements, with the full five-action set $\{a_1,a_2,a_3,a_4,a_5\}$ achieving optimal results (64.7 average). Notably, even minimal subsets demonstrate reasonable performance, with the single-action baseline $\{a_5\}$ (equivalent to standard Chain-of-Thought) achieving 55.9 average accuracy. The incremental gains from $\{a_5\}$ to $\{a_1, a_5\}$ (+3.2 points) and subsequent expansions demonstrate that each additional action contributes meaningfully to the reasoning process. These results highlight two key insights: (1) TemplateRL's robustness to action space configuration, enabling deployment in resource-constrained scenarios with reduced action sets, and (2) the complementary nature of different reasoning actions, where diverse problem-solving strategies collectively enhance performance. This flexibility allows practitioners to adapt the framework based on task-specific requirements.

\subsection{Sensitivity to Training Data Size}\label{D.5}
As mentioned in the main paper, we select training data exclusively from the MATH dataset~\citep{hendrycks2021measuring}, using only 5k MATH level 3-5 problems for RL training. Given the prevalence of large-scale datasets, we investigate whether there exists a scaling law for performance versus training data size. Specifically, we augment our original 5K sample set by leveraging the OpenAI-o1 model to generate an additional 10K synthetic problems of similar difficulty, along with their solutions. To ensure solution quality, we employ deepseek-R1 for further verification and correction, filtering out erroneous samples and refining incorrect solutions. This process yields 10k high-quality synthetic data that maintains consistency with the original MATH distribution. We then conduct experiments using mixed datasets of varying sizes: 5k (original), 10k (original + 5k synthetic), and 15k (original + 10k synthetic).

As shown in Table~\ref{training_data_size}, TemplateRL exhibits positive scaling behavior with increased training data. Performance improves from 64.7\% (5K) to 66.1\% (10K) and further to 67.1\% (15K), representing consistent gains of +1.4 and +2.4 percentage points respectively. Notably, the most substantial improvements occur on the challenging AIME24 benchmark (+3.4 percentage points from 5K to 10K), suggesting that additional training data particularly benefits complex reasoning tasks. While the scaling trend is encouraging, we observe that TemplateRL can achieve competitive performance even with limited training data, making it practical for resource-constrained scenarios. More comprehensive investigations into sample diversity and scaling to larger datasets are left for future work.

\begin{table}[t!]
\resizebox{1.0\linewidth}{!}{
\centering
\begin{tabular}{lcccc} 
    \toprule
    Data Size & MATH500~$\uparrow$ & AIME24~$\uparrow$ & AMC~$\uparrow$ & Avg.~$\uparrow$ \\
    \midrule
    \rowcolor{pink!15}
    5K (Original) & 83.4 & 33.3 & 77.5 & 64.7\\
    10K (+ 5K Syn.) & 84.0 & 36.7 & 77.5 & 66.1\\
    15K (+ 10K Syn.) & 84.6 & 36.7 & 80.0 & 67.1\\
    \bottomrule
\end{tabular}}
\caption{Performance scaling with training data size on Qwen2.5-Math-7B-Base. `Syn.' represents synthetic data. Results demonstrate positive scaling behavior, with notable improvements on challenging benchmarks.}
\label{training_data_size}
\end{table}

\begin{table}[t!]
\resizebox{1.0\linewidth}{!}{
\centering
\begin{tabular}{lcccc} 
    \toprule
    Metric & MATH500~$\uparrow$ & AIME24~$\uparrow$ & AMC~$\uparrow$ & Avg.~$\uparrow$ \\
    \midrule
    SC-only & 81.2 & 30.0 & 72.5 & 61.2\\
    PS-only & 80.8 & 26.7 & 72.5 & 60.0\\
    \rowcolor[RGB]{236,244,252}
    PCC-only & 83.4 & 33.3 & 77.5 & 64.7\\
    \rowcolor{pink!15}
    Unified (Equal) & 83.9 & 36.7 & 77.5 & 66.0\\
    \bottomrule
\end{tabular}}
\caption{Performance comparison across different pattern-matching metrics on Qwen2.5-Math-7B-Base. ``Equal'' denotes equal weights for three indicators.}
\label{pattern-match}
\end{table}

\subsection{Sensitivity to Pattern-Matching Metrics}\label{D.6}
Beyond the Problem Condition Complexity (PCC) metric~\citep{lee2000problem, SALADO2014539} 
employed in the main paper for retrieving optimal thought patterns, we explore 
alternative pattern-matching strategies to assess the robustness of our approach. 

\paragraph{Metric Design Rationale.}
Our choice of PCC is motivated by its structural focus: unlike semantic embeddings 
that primarily capture surface-level similarity in problem statements, PCC quantifies 
the \emph{structural complexity} of problems based on the number of prior conditions 
and constraints. This structural feature directly informs the appropriate reasoning 
strategy—high-PCC problems often require divide-and-conquer approaches to manage 
multiple interdependent conditions, while simpler low-PCC problems may need only 
direct reasoning. This explicit complexity-strategy mapping enhances both 
interpretability (templates are matched based on observable problem structure) 
and robustness (matching is less sensitive to superficial phrasing variations).

\paragraph{Alternative Metrics Comparison.}
To empirically validate PCC's effectiveness, we investigate three additional 
pattern-matching strategies: 
\begin{itemize}[leftmargin=*,itemsep=0pt]
    \item \textbf{Subquestion-count (SC)}: The number of subquestions an 
          unstructured problem can be decomposed into, reflecting reasoning 
          granularity;
    \item \textbf{Problem semantics (PS)}: Semantic similarity-based matching 
          using sentence embeddings (all-MiniLM-L6-v2), representing surface-level 
          textual similarity;
    \item \textbf{Unified ranking (UR)}: A composite metric that computes PCC, 
          SC, and PS for each incoming question, performs triple relevance ranking 
          across thought templates, and selects top candidates through weighted 
          averaging ($w_{PCC}$=0.5, $w_{SC}$=0.3, $w_{PS}$=0.2).
\end{itemize}

As shown in Table~\ref{pattern-match}, while the unified ranking metric achieves 
the highest overall performance (66.0 average), the PCC-only approach demonstrates 
competitive results (64.7 average) with significantly lower computational overhead. 
Notably, semantic-based matching (PS-only) shows the weakest performance (60.2 
average), which aligns with our hypothesis: mathematical reasoning patterns are 
inherently structural rather than semantic—two problems may share similar wordings 
yet require entirely different reasoning strategies, while problems with distinct 
phrasings may share the same underlying structural complexity. Surface-level 
semantic similarity fails to capture these critical strategic distinctions, whereas 
PCC's structural focus naturally aligns template matching with the actual reasoning 
demands of problems.

Given that PCC provides a simple, interpretable, and effective solution with 
minimal computational cost, we adopt it as our default pattern-matching metric. 
The strong performance of the unified approach (UR) suggests that combining 
multiple metrics can further enhance matching quality, and future work could 
explore more sophisticated weighting schemes or learned combination functions 
to optimize this trade-off between performance and efficiency.

\begin{table}[t!]
\centering
\resizebox{1.0\linewidth}{!}{
\begin{tabular}{ccccc} 
    \toprule
    Seed Data & MATH500 & AIME24~$\uparrow$ & AMC~$\uparrow$ & Avg.~$\uparrow$ \\
    \midrule
    100 & 77.4 & 20.0 & 65.0 & 54.1 \\
    200 & 81.0 & 26.7 & 72.5 & 60.0 \\
    \rowcolor{pink!15}
    500 & 83.4 & 33.3 & 77.5 & 64.7 \\
    1000 & 85.2 & 36.7 & 77.5 & 66.5 \\
    \bottomrule
\end{tabular}}
\caption{Impact of seed data size.}
\label{table_sensitivity}
% \vskip -0.1in
\end{table}

\begin{table*}[t!]
% \resizebox{1.0\linewidth}{!}{
\centering
\begin{tabular}{cccccc} 
   \toprule
   Action  & DC ($a_1$) & SR ($a_2$) & SA ($a_3$) & OST ($a_4$) & CoT ($a_5$)\\
   \midrule
   Ratio & 31.3\% & 7.2\% & 5.7\% & 30.3\% & 25.5\%\\
   \bottomrule
\end{tabular}
% }
\caption{Statistics of reasoning actions: proportion of each action relative to all actions in the template library.}
\label{action statistics}
\end{table*}

\subsection{Sensitivity to Seed Data Size.}\label{seed data}
As shown in Table~\ref{table_sensitivity}, performance scales with seed data size: average accuracy improves from 38.7\% (100 samples) to 51.4\% (1,000 samples). Notably, TemplateRL achieves competitive performance even with 100 samples, highlighting practicality in resource-constrained settings. We use 500 samples by default, balancing performance enhancement and data efficiency.

\subsection{Statistics on Reasoning Actions}\label{D.8}
We present the occurrence statistics of different reasoning actions in the constructed thought template library in Table~\ref{action statistics}. For brevity, DC, SR, SA, OST, and CoT denote \textsc{Divide\_and\_Conquer}, \textsc{Self\_Reflection}, \textsc{System\_Analysis}, \textsc{One\_Step\_Thought}, and \textsc{Chain\_of\_Thought}, respectively. The results reveal a relatively balanced distribution across actions, with DC and OST demonstrating the highest frequencies (31.4\% and 30.3\%), followed by CoT (25.5\%), while SR and SA show lower occurrences (7.2\% and 5.7\%). This distribution pattern indicates that during MCTS exploration, OST, DC, and CoT are more effective at generating correct solutions and resulting in their elevated ratio within the thought template library.

\subsection{Statistical Analysis}\label{D.9}
To statistically evaluate the significance of performance differences between TemplateRL and baseline methods, we apply the nonparametric Wilcoxon signed-rank test~\citep{wilcoxon1992individual}. This statistical test is specifically designed to compare paired observations when data may not follow a normal distribution, making it particularly suitable for performance analysis across multiple benchmarks. The Wilcoxon signed-rank test evaluates whether there is a significant difference between paired observations through the following procedure:
\begin{enumerate}
\item \textbf{Calculate differences:} For each benchmark pair, compute the difference $D_i = X_i - Y_i$ where $X_i$ represents TemplateRL performance and $Y_i$ represents baseline performance.
\item \textbf{Rank differences:} Take absolute values $|D_i|$ and rank them from smallest to largest as $R_i$, with average ranks assigned for ties.
\item \textbf{Assign signs to ranks:} For each difference $D_i$, assign its sign to the corresponding rank: $R'_i = \text{sign}(D_i) \cdot R_i$.
\item \textbf{Calculate rank sums:} Compute positive and negative rank sums: $W^+ = \sum_{D_i > 0} R'_i$ and $W^- = \sum_{D_i < 0} R'_i$.
\item \textbf{Determine test statistic:} The test statistic is $W = \min(W^+, W^-)$.
\item \textbf{Calculate p-value:} Derive the p-value from the distribution of test statistic $W$.
\end{enumerate}

We test the null hypothesis $H_0$: no significant difference between methods against the alternative hypothesis $H_1$: significant difference exists. Comparing TemplateRL with GRPO on Qwen2.5-Math-7B-Base across all benchmarks, we obtain a p-value of 0.0156. Using a significance level of $\alpha = 0.05$, we reject the null hypothesis ($p < 0.05$), providing strong statistical evidence that TemplateRL significantly enhances reasoning performance compared to standard GRPO training.

\section{Case Study}\label{E}
\begin{figure*}[htbp!]
\begin{center}
% \vskip 0.2in
\centerline{\includegraphics[width=0.92\textwidth]{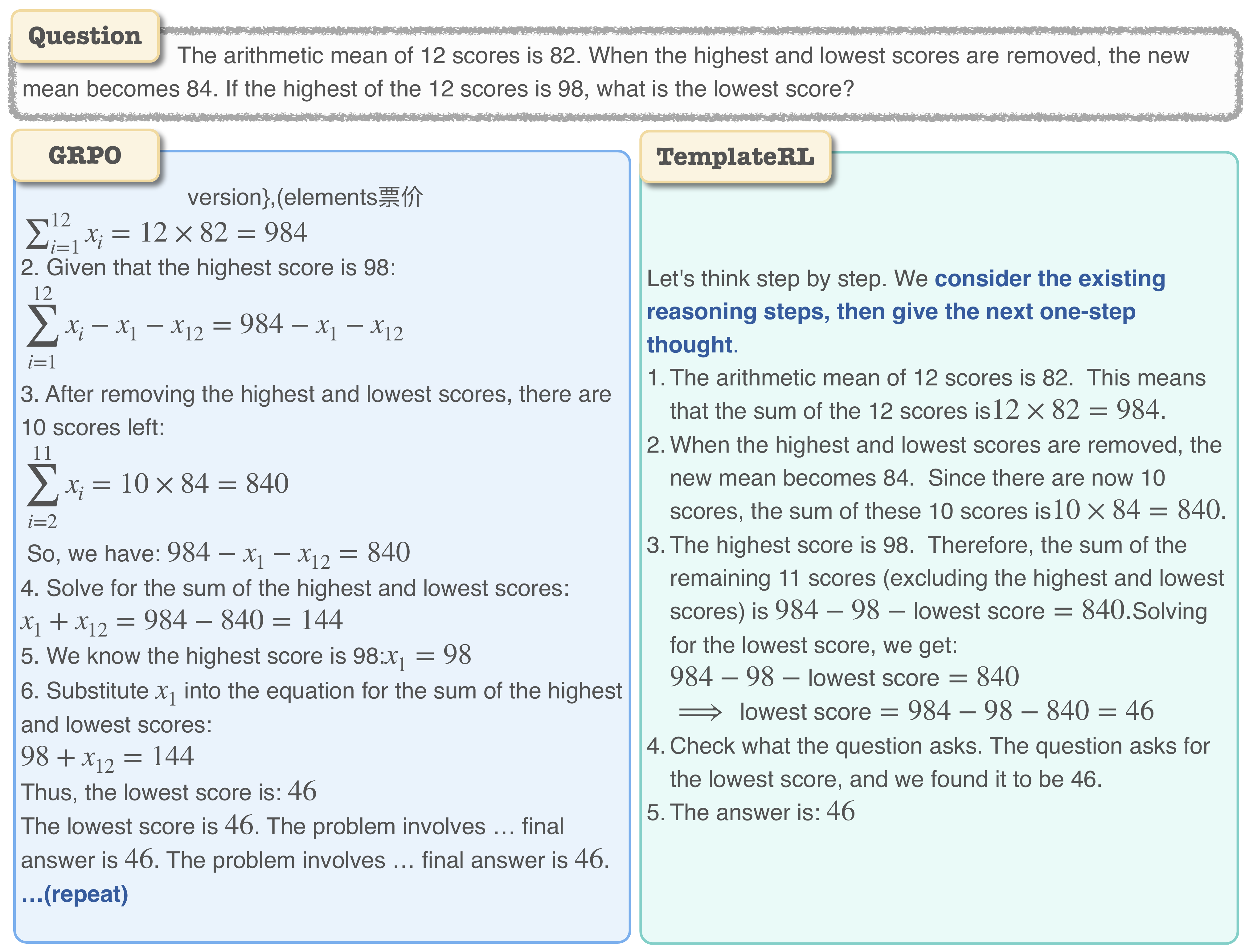}}
% \vskip 0.1in
\caption{Comparison of GRPO and TemplateRL for a simple algorithm problem from the MATH dataset.}
\label{case1}
\end{center}
% \vskip 0.15in
\end{figure*}

\begin{figure*}[ht!]
\begin{center}
% \vskip 0.1in
\centerline{\includegraphics[width=0.95\textwidth]{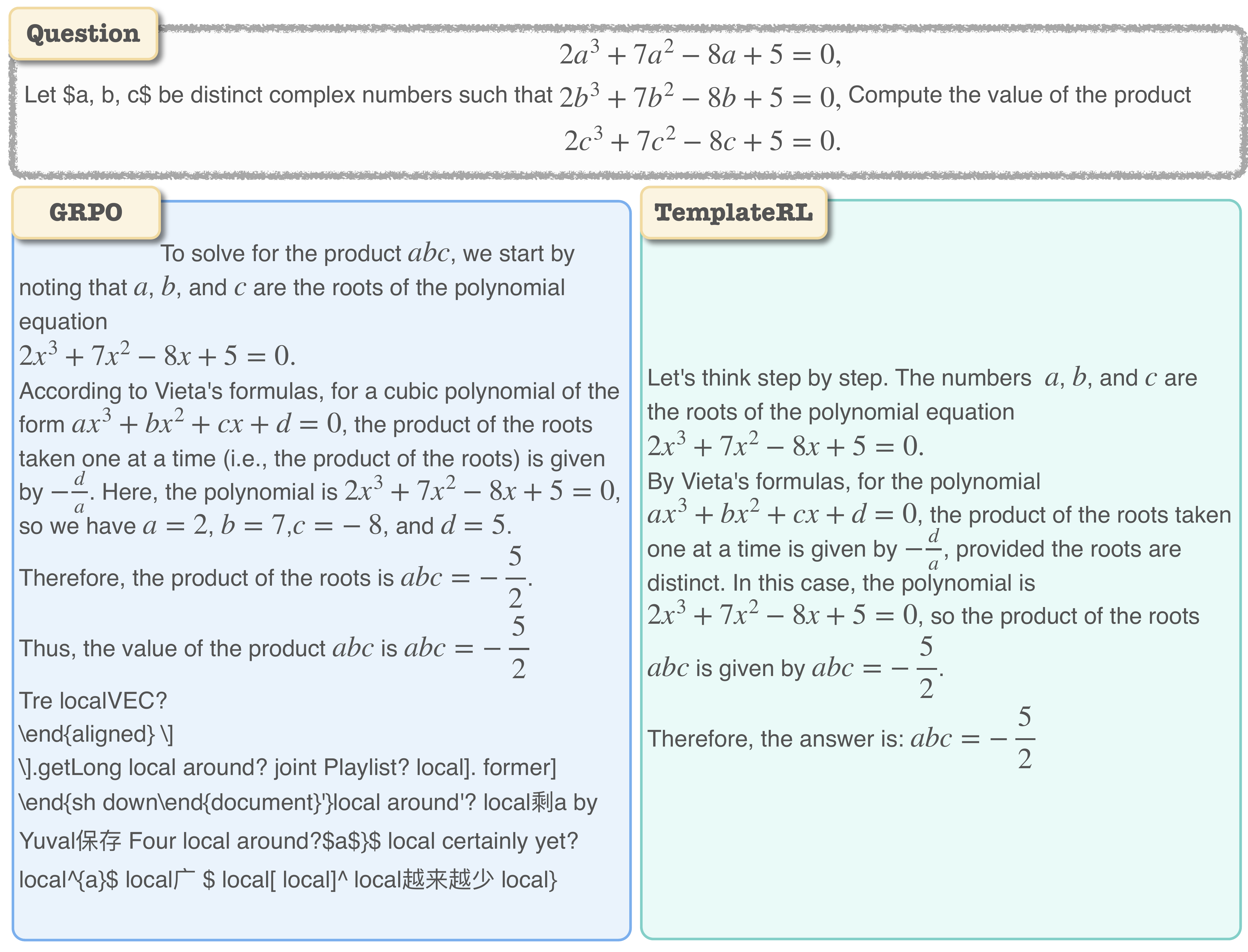}}
% \vskip 0.1in
\caption{Comparison of GRPO and TemplateRL for a difficult algorithm problem from the MATH dataset.}
\label{case2}
\end{center}
% \vskip 0.15in
\end{figure*}

To further analyze the improvements of TemplateRL over conventional GRPO, we compare their reasoning processes on representative mathematical problems from the MATH dataset in Figure~\ref{case1} and Figure~\ref{case2}.

In Figure~\ref{case1}, we observe how both methods approach an arithmetic mean problem. GRPO produces a solution with scattered notation and repetitive statements, particularly evident in its final steps where it repeatedly states ``The problem involves ... final answer is 46.'' In contrast, TemplateRL demonstrates a more structured approach by explicitly introducing a ``step by step'' thinking process. The thought-augmented process methodically builds upon each reasoning step, clearly identifying the relationship between the original sum (984), the highest score (98), and the remaining scores to derive the lowest score (46). This structured approach results in a more readable and logically coherent solution.

Figure~\ref{case2} presents a more challenging problem involving complex numbers and Vieta's formulas. Here, the limitations of GRPO become more pronounced. While GRPO initially applies the correct formula, its reasoning process deteriorates into incoherent text fragments and coding artifacts (e.g., ``Tre localVEC?'' and various non-mathematical expressions). This demonstrates how GRPO struggles with maintaining coherent reasoning for complex problems. In contrast, TemplateRL maintains its structured approach throughout, clearly stating the problem context, applying Vieta's formulas with proper explanation, and presenting a clean, coherent solution without extraneous text or errors.

\section{Additional Discussion}\label{F}
\subsection{Justification for Choosing MCTS.}
The selection of MCTS is driven by both task-specific requirements and its comparison with other search algorithms.

\paragraph{Task Requirements:}
Our task involves constructing complex reasoning chains from a vast combinatorial action space. This setting demands an algorithm that can efficiently search and optimize within such a large space. MCTS is particularly well-suited for this, as it incrementally explores the most promising trajectories while maintaining flexibility through stochastic sampling.

\paragraph{Limitations of Other Search Methods:}
We investigated alternatives such as Breadth-First Search (BFS) and Depth-First Search (DFS), and identified several issues:
\begin{itemize}
    \item DFS is vulnerable to early-stage errors. A suboptimal initial action can lead the search deep into unproductive paths, making recovery and backtracking difficult in multi-step reasoning where early decisions are often uncertain.
    \item BFS is computationally expensive, exploring the reasoning space level by level with exponential growth in candidate states, making it intractable for practical problems.
\end{itemize}

\paragraph{MCTS Advantages:}
MCTS offers a superior balance between exploration and exploitation:
\begin{itemize}
    \item Compared to DFS, MCTS does not commit to a single path. It uses rollout and backpropagation to update the estimated value of nodes, allowing the search to adaptively shift away from poor paths.
    \item Compared to BFS, MCTS grows the search tree asymmetrically, guided by UCT (Upper Confidence Bound for Trees). This selective expansion enables it to focus computation on the most promising subtrees, significantly improving efficiency.
\end{itemize}

\paragraph{Computational Efficiency:}
Importantly, MCTS template construction incurs negligible computational overhead in our framework. On 8×A100-80GB GPUs, constructing the template library from 500 seed samples requires about \textit{20 minutes}, compared to the \textit{$\sim$10-20 hours} required for full RL training on 5K samples (500 steps). Moreover, this is a one-time cost: once constructed, the template library can be reused across different model scales, architectures, and even domains (as demonstrated by our cross-domain transfer results). In most scenarios, 500 seed samples are sufficient for building the initial library.

In summary, given its ability to handle large search spaces, its robustness to early decision errors, its structural alignment with sequential reasoning, and its minimal computational cost, MCTS emerged as an effective and efficient approach for our reasoning framework.

\vspace{-0.05in}
\subsection{Justification of Key Design Choices}\label{sec:design_justification}
We justify two critical decisions:

\textbf{(1) Five-action space:} The action set \{DC, SR, SA, OST, CoT\} was designed based on cognitive science literature~\citep{Kahneman2011} and validated through pilot studies. Testing alternatives on 500-sample validation (diverse task samples) showed: minimal 2-action set (62.4\% average accuracy), our 5-action design (71.3\%), extended 8-action set (72.4\%, only +1.1). The design principle is domain-agnostic meta-strategies rather than domain-specific operations.

\textbf{(2) PCC over semantic similarity:} PCC (Problem Condition Complexity) correctly distinguishes problems by reasoning requirements rather than surface similarity. For example, ``find $x$ where $x^2+3x-4=0$'' (PCC=1) vs. ``find $x,y$ where $x+y=5$ and $xy=6$'' (PCC=2) appear semantically similar but require different strategies (direct CoT vs. DC→CoT).

\subsection{Why A Small Set of Generic Patterns Can Generalize?}
Our templates encode high-level strategic patterns rather than problem-specific solutions. These templates capture abstract reasoning strategies such as ``decompose → solve subproblems → verify'' that operate above the surface-level content. While individual problems vary widely in their mathematical details, the underlying reasoning structures often follow recognizable strategic patterns, which is analogous to how humans apply divide-and-conquer or proof-by-contradiction across diverse domains. We believe that not only can individual actions generalize, but their combinations can as well, particularly for complex real-world scenarios where solutions emerge from abstracting more sophisticated action sequences.

\subsection{Are Templates Optimized for Test Sets?}
No. As shown in Figure~\ref{ood_figure}, templates constructed from mathematical problems generalize effectively to diverse domains: BALROG (+6.1\% performance gains for agentic reasoning), GPQA-D (+6.6\% gains for graduate-level science), and MMLU-Pro (+7.4\% gains for general knowledge). These substantial cross-domain improvements demonstrate that templates are general, not test-set optimized. These templates capture transferable strategic patterns rather than domain-specific solutions.

\subsection{How does performance scale with template set size?}
In our preliminary experiments, we explored the impact of varying template set sizes and observed a clear pattern: performance improves substantially when increasing template set size below 100 templates, but gains begin to slow beyond this threshold. This suggests an optimal balance point where the template library is sufficiently diverse to cover common reasoning patterns without introducing retrieval complexity or redundancy. Future work could investigate adaptive mechanisms that dynamically adjust the template set based on problem domain or difficulty.

\subsection{Robustness to Seed Set and Cross-Domain Generalizability}
A potential concern is whether the template library is overly dependent on the quality or domain coverage of the initial 500 seed samples. However, our empirical results demonstrate that the high-level thought patterns abstracted via MCTS possess remarkable task-agnostic properties. As shown in Figure \ref{ood_figure}, templates derived exclusively from mathematical seeds significantly enhance performance in entirely different domains, such as agentic embodied reasoning (BALROG, +6.1\%) and graduate-level science (GPQA-D, +6.6\%). 

This indicates that TemplateRL captures universal ``System 2'' strategic logic—including problem decomposition, iterative self-reflection, and constraint analysis—rather than surface-level heuristics. 
The high-level nature of these patterns mitigates the risk of the model being misled by specific seed biases into suboptimal paths. 
Therefore, in practical applications, selecting a seed set that covers a wide range of difficulty levels is sufficient to build a robust template library that facilitates effective policy optimization across diverse tasks.

\subsection{Distinction between TemplateRL and Related Works}
A related line of work, ReMA~\cite{wan2025rema}, learns meta-cognitive strategies through multi-agent reinforcement learning, where agents discover abstract reasoning operators via agent-level reward signals. While ReMA focuses on \emph{implicit} meta-strategy discovery through agent interactions, TemplateRL constructs \emph{explicit}, interpretable template libraries via systematic MCTS exploration, enabling immediate human inspection, manual editing, and efficient cross-domain transfer without the multi-agent training overhead.

\subsection{Future Work}
In future work, we plan to extend TemplateRL along several directions. First, we aim to scale the framework to more complex agentic scenarios~\cite{tang2025agent,chen2025beyond,wu2026spark,xu2026odysseyarena}, where dynamic branching and multi-step decision-making demand richer template structures beyond the current five-action space. We also hope to incorporate the entropy mechanism into training to further enhance template diversity~\cite{kang2025entropy}. Second, we will explore integrating TemplateRL with heterogeneous model orchestration~\cite{wu2026atlas}, enabling structured template guidance across multi-domain reasoning tasks that require coordinating specialized tools and models. Additionally, we aim to extend templates to more general domains, such as dialogue, retrieval-augmented generation, role-playing, and optimize templates for more diverse objectives~\cite{chen-etal-2024-retrieval-style,wu-etal-2025-pandoras,liao2025moa}. When deployed in multi-agent settings, we further plan to investigate how communication topologies influence the propagation of structured templates across agents~\cite{shen2025understanding}, and enhance the robustness of template-guided coordination against unknown adversarial conditions~\cite{miao2025blindguard}. Third, to improve deployment efficiency, we plan to combine TemplateRL with model compression~\cite{feng2025two} and inference acceleration techniques~\cite{shen2026double}, reducing the computational footprint of structured rollout generation while preserving the performance gains demonstrated in this work.

\setcounter{tocdepth}{-1}
\addtocontents{toc}{\protect\setcounter{tocdepth}{2}}

\end{document}